\documentclass[11pt, authoryear]{article}

\usepackage[utf8]{inputenc}
\usepackage[export]{adjustbox}
\usepackage[ruled,vlined, linesnumbered]{algorithm2e}
\usepackage{algpseudocode}
\usepackage{amsmath}
\usepackage{amssymb}
\usepackage{animate}
\usepackage{array}
\usepackage{authblk}
\usepackage{bm}
\usepackage{booktabs}
\usepackage{caption}
\usepackage{capt-of}
\usepackage{color}
\usepackage{csvsimple}
\usepackage{enumitem}
\usepackage{fontawesome5}
\usepackage[T1]{fontenc}
\usepackage{framed}
\usepackage{graphicx}
\usepackage[colorlinks=true, linkcolor = black, citecolor = black, urlcolor = blue]{hyperref}
\usepackage[utf8]{inputenc}
\usepackage{longtable}
\usepackage{lscape}
\usepackage{makecell}
\usepackage{multicol}
\usepackage{multirow}
\usepackage[numbers, round, sort&compress]{natbib}
\usepackage{realboxes}
\usepackage{rotating}
\usepackage{setspace}
\usepackage{soul}
\usepackage{subcaption}
\usepackage{svg}
\usepackage[most]{tcolorbox}
\tcbuselibrary{listings}
\usepackage{url}
\usepackage{xcolor}

\graphicspath{ {Figures/} }
\newcolumntype{C}[1]{>{\centering\arraybackslash}m{#1}}
\newcommand{\instructions}[1]{}

\definecolor{light-gray}{gray}{0.96} 
\definecolor{dark-gray}{gray}{0.8} 

\DeclareRobustCommand{\hldgray}[1]{{\sethlcolor{dark-gray}\hl{#1}}}

\lstnewenvironment{VerbItem}{%
    \lstset{basicstyle=\ttfamily\small,breakatwhitespace,breaklines,
            numbers=left,numbersep=2pt,xleftmargin=0.75em}}{}
\usepackage{authblk}
\usepackage{ulem}
\usepackage{listings} 
\usepackage{comment}

\usepackage{arxiv}
\usepackage{doi}

\title{How Generative AI models such as ChatGPT can be (Mis)Used in SPC Practice, Education, and Research? An Exploratory Study}
\renewcommand{\shorttitle}{ChatGPT and SPC}

\author[1]{Fadel M. Megahed}
\author[2]{Ying-Ju Chen}
\author[1]{Joshua A. Ferris}
\author[3]{Sven Knoth}
\author[1,*]{L. Allison Jones-Farmer}

\affil[1]{Farmer School of Business, Miami University, Oxford, OH 45056, USA}
\affil[2]{Department of Mathematics, University of Dayton, OH 45469, USA}
\affil[3]{Department of Mathematics \& Statistics, Helmut Schmidt University, Hamburg, Germany}
\affil[*]{Corresponding author. Can be reached at \href{mailto:farmerl2@miamioh.edu}{farmerl2@miamioh.edu}}

\date{February 16, 2023}

\begin{document}

\maketitle

\bigskip

\begin{abstract}
\noindent Generative Artificial Intelligence (AI) models such as OpenAI's ChatGPT have the potential to revolutionize Statistical Process Control (SPC) practice, learning, and research.  However, these tools are in the early stages of development and can be easily misused or misunderstood. In this paper, we give an overview of the development of Generative AI. Specifically, we explore ChatGPT's ability to provide code, explain basic concepts, and create knowledge related to SPC practice, learning, and research. By investigating responses to structured prompts, we highlight the benefits and limitations of the results. Our study indicates that the current version of ChatGPT performs well for structured tasks, such as translating code from one language to another and explaining well-known concepts but struggles with more nuanced tasks, such as explaining less widely known terms and creating code from scratch.  We find that using new AI tools may help practitioners, educators, and researchers to be more efficient and productive. However, in their current stages of development, some results are misleading and wrong. Overall, the use of generative AI models in SPC must be properly validated and used in conjunction with other methods to ensure accurate results.  
\end{abstract}

\vspace{\baselineskip}


\clearpage


\section{Preface}\label{sec:preface}

``What do you think of ChatGPT?'' Many readers of this paper have asked and/or been asked this question, likely sometime between December 2022 and January 2023. Writing an article about large language models (LLM), such as ChatGPT, is a daunting task since: (a) the technology is rapidly evolving; (b) the content can cover many aspects, including the details of a given model and ethical/legal considerations about the role of artificial intelligence (AI) in our society; and (c) it is hard to forecast the ubiquitousness of such tools in a few years. In this paper, we focus on what LLM models can and cannot do well \ul{now}. 

The overarching objective of this expository paper is 
to examine whether LLMs such as ChatGPT-like models \textbf{can be useful in statistical process control (SPC) applications}. Our examination will focus on three primary SPC application domains: \textbf{\ul{practice}}, \textbf{\ul{learning/training}}, and \textbf{\ul{research}}. We do not discuss important issues such as ethical, legal, and other philosophical concerns from using AI tools.

\section{Introduction}\label{sec:intro}

The rapid developments in AI over the past three years have changed the general perception of what AI can do. In the SPC community, we have equated AI to predictive machine learning (ML) systems that are trained to solve classification, regression, or clustering problems \citep{megahed2015statistical,weese2016statistical,colosimo2021artificial}. While ML remains an essential component of AI, state-of-the-art models have transitioned to ``generative AI'' where the objective is to generate new content rather than analyze an existing dataset \citep{gozalo2023chatgpt}. The generated content is based on a stochastic behavior embedded in generative AI models such that the same input prompts results in different content. 

State-of-the-art generative AI models can have up to 175 billion parameters. 
With the increase in model size, researchers have observed the ``emergent abilities'' of LLMs, which were not explicitly encoded in the training \citep{wei2022emergent}. Examples of ~``prompted tasks with emergent performance [included]: multi-step arithmetic, taking college-level exams and identifying the intended meaning of a word.'' \citep{wei2022characterizing}

These models have allowed a wide range of applications. For example, GPT-3 has been used in language translation, text summarization, and content generation applications \citep{devlin2018bert}. To reduce the level of expertise needed to use and deploy those generative AI models, AI-powered chatbots and virtual assistants, such as ChatGPT and CoPilot, have been developed \citep{mckee2022chatbots}. For example, ChatGPT is a chatbot-like variation of GPT-3 (technically GPT-3.5) that has been fine-tuned for conversational language understanding and generation, allowing it to respond more human-like in a conversational context. In addition to text-to-text generative AI tools, several breakthroughs in text-to-audio (e.g., \textit{Google's} MusicLM \citep{agostinelli2023musiclm}), text-to-image (e.g., \textit{OpenAI's} DALL.E 2 \citep{openai2022dalle2}), and text-to-video (e.g., \textit{Meta's} Make-A-Video \citep{singer2022make} and Make-A-Video3D \citep{singer2023make}) applications have been made. 

\begin{figure}[htb]
    \centering
    \begin{framed}
    \begin{subfigure}[b]{\textwidth}
        \centering
        \includegraphics[width=\textwidth]{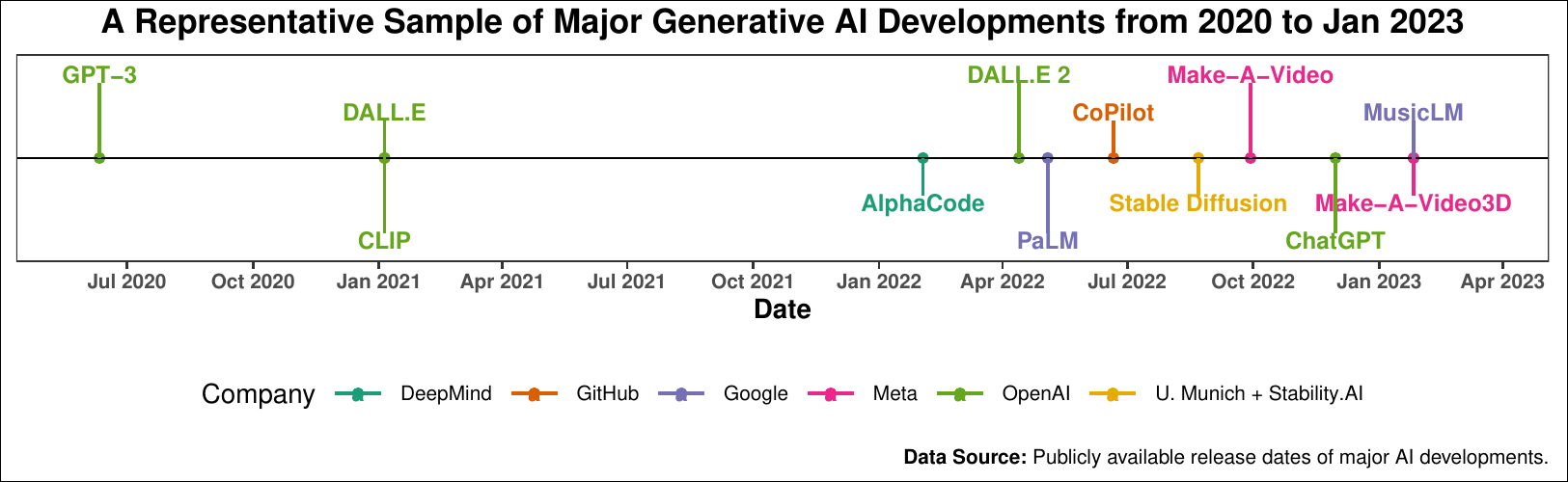}
        \caption{A representative sample of LLM models released/published in the last three years.}
        \label{fig:LLM_models}
    \end{subfigure}

    \vspace{\baselineskip}
    
   \begin{subfigure}[b]{\textwidth}
        \centering
        \includegraphics[width=\textwidth]{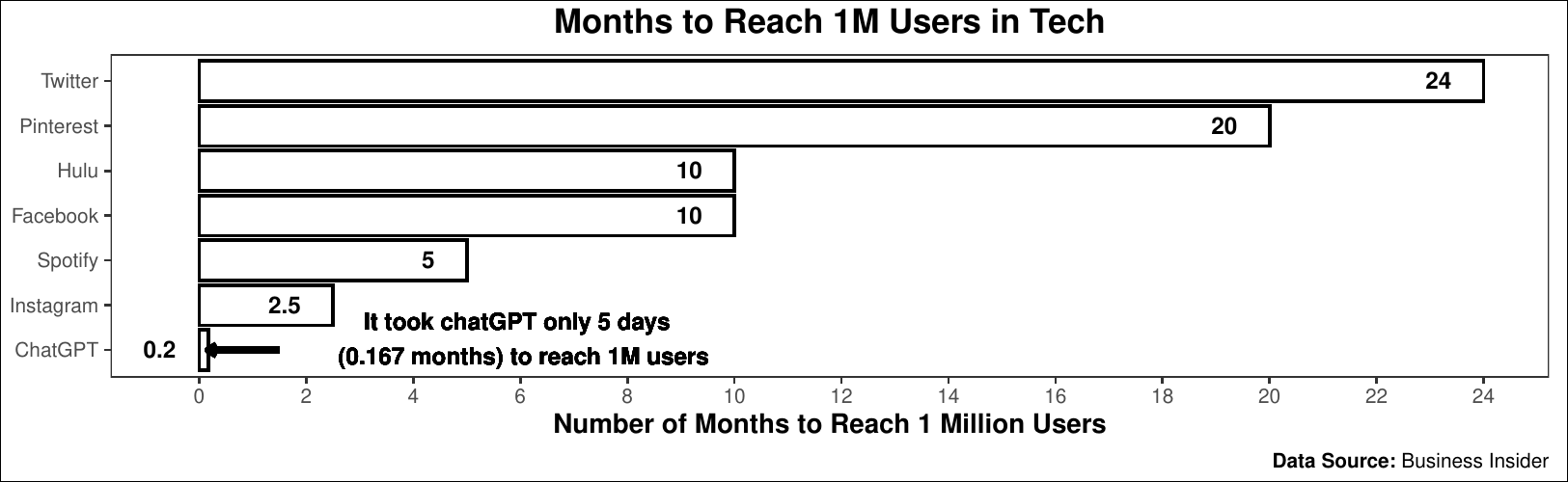}
        \caption{The unprecedented adoption of ChatGPT compared to popular technology platforms.}
        \label{fig:chatGPT}
    \end{subfigure}
    \end{framed}

    \vspace{-\baselineskip}
    
    \caption{The rapid developments in generative AI and the associated hype in trying/adopting these tools by the general public.}
    \label{fig:introductory}
\end{figure}

The recent generative AI breakthroughs (see Figure \ref{fig:LLM_models}) and adoption of these tools (as in ChatGPT's case depicted in Figure \ref{fig:chatGPT}) have created hype in the business community. \textit{Sequoia Capital} estimates that ``generative  AI can make [knowledge and creative] workers at least 10\% more efficient and/or creative: they become not only faster and more efficient, but more capable than before. Therefore, Generative AI has the potential to generate trillions of dollars of economic value.'' \citep{huang2023generative} Their viewpoint reflects the broader investor community's sentiment as it invested \$1.37 billion in generative AI in 2022, approximately equal to their combined investment in the previous five years \citep{griffith2023-new}. 

In this article, we ask, ``what can generative LLM-based AI tools do now to augment the roles of SPC practitioners, educators, and researchers?''  To make our task more tractable, we will primarily focus on evaluating the utility of ChatGPT (and its underlying GPT-3.5 engine) since it: (a) is the most well-known of these generative AI tools and (b) combines features of the generative chatbot with an underlying LLM that can generate both text and code. 
In our estimation, this expository assessment can provide a benchmark for future evaluations of the next generation of generative AI models. We hope this study motivates SPC practitioners and researchers to be receptive to exploring whether new AI tools can help them be more efficient, productive, and innovative. This is consistent with the recommendations of (a) \citet[p. 20]{box2012innovation} who stated that ``we stress the necessity for the quality engineering community to strengthen and promote its role in innovation'', and (b) \citet[p. 165]{hockman2016statisticians} who stated that ``for statisticians to be successful in leading innovation, they will need to strengthen their skills beyond what they have traditionally needed in the past, but we believe this will be worth the effort''.


\section{Background}\label{sec:back}


The notion of a language model has risen in popularity in the last decade thanks to modeling techniques and computational power advances. Language models are used to assign probabilities to word sequences. Interestingly, the foundation of language models was set by the mathematician Andrey A. Markov in the early 1900s, who used Alexander Pushkin's Eugene Onegin novel to demonstrate that letter pair sequences are not independent and that the likelihood of a letter's appearance can be approximated \citep{markov2006}. \cite{shannon1948language} built on that work, developing a statistical model for the English language and showing that text sequences could be generated. The work of Markov and Shannon is the foundation of modern language work; modern models will use sequences of words rather than events (Markov) or characters/words (Shannon). Today, language models are ubiquitous. Examples include text auto-completion (used in web browsers, mobile phones' messaging apps, software GUIs, etc.), machine translation, natural language generation, and optical character recognition.

The development of the \textit{Transformer} model set the foundation for today's large language models. The \textit{Transformer} model was proposed in \cite{vaswani2017attention} as a deep learning model grounded on self-attention mechanisms. The self-attention mechanism assigns different weights according to the importance of each part of the data. The \textit{Transformer} model architecture allows for parallel computing, thus reducing training time \citep{vaswani2017attention}. The parallelization advantage and the model's architecture result in better performance and accuracy than traditional recurrent neural network-based deep learning models. Hence, existing LLMs utilize a \textit{Transformer-}based architecture, making it possible to train multi-billion parameter models on several terabytes of text.

OpenAI, an AI research lab, introduced its third-generation autoregressive language model, \textit{Generative Pre-trained Transformer} 3 (GPT-3), which employs deep learning to generate human-like text in 2020 \citep{brown2020language}. GPT-3 was pre-trained on a diverse corpus of unlabeled contiguous text containing almost 500 billion tokens, and the model had 175 billion parameters making it, at the time, the largest natural language processing (NLP) model \citep{radford2018improving}. GPT-3 learns patterns and relationships in the data without a label, meaning the model will have no understanding of fact or fiction, just the patterns of words. During training, it predicts the next word in a text string, given the previous words as input. The model's parameters are optimized using the gradient descent algorithm, which improves its ability to generate text that resembles the text in the training data over time. Once trained, the model can be refined by adding task-specific data to train the model for specific NLP tasks such as text generation, summarization, translation, and question answering. By combining pre-training on a large corpus of diverse text data and fine-tuning on task-specific data, GPT-3 can perform various NLP tasks with high fidelity.


GPT-3 has proven to be a powerful tool for natural language, \cite{brown2020language} noted in their paper the potential to apply the model to programming applications. OpenAI created the Codex model, a fine-tuned GPT-3 for programming tasks with 12 billion parameters, and trained using select files from 54 million public GitHub repositories \citep{chen2021evaluating}. Codex powers GitHub Copilot, a cloud-based tool embedded into several Integrated Development Environments (IDEs) such as Visual Studio, Visual Studio Code, JetBrains, and Neovim.

\begin{figure}[htbp]
  \centering
  \begin{framed}
  \includesvg[width=0.94\textwidth]{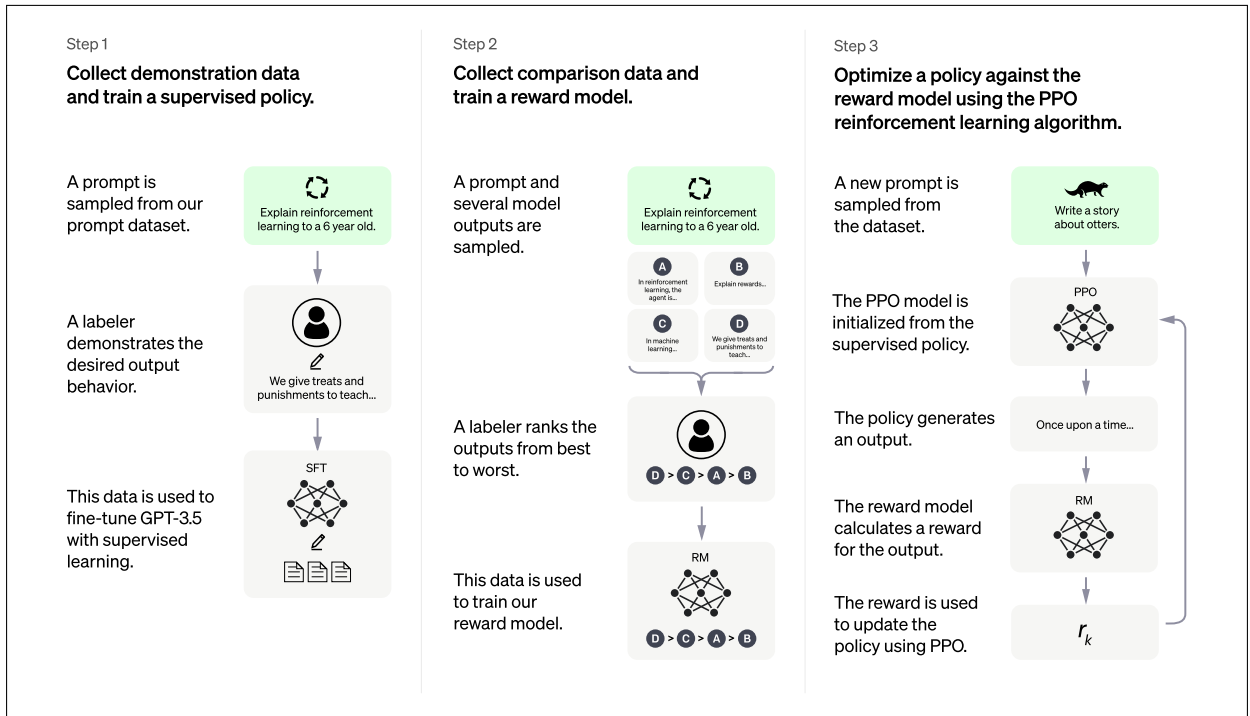}
  \end{framed}
  \vspace{-\baselineskip}
  \caption{The ChatGPT training process. The figure is from \citet{openai2022chatgpt}.}\label{fig:ChatGPT_Diagram}
\end{figure}

ChatGPT is a pre-trained instance of GPT-3, specifically optimized for the task of generating conversational responses.  
While GPT-3 can be used for various NLP tasks, ChatGPT is designed to generate human-like responses to the text input conversationally. Over time, the model learns the patterns and relationships in the data, improving its ability to generate text that resembles the text in the training data. This mechanism allows ChatGPT to generate consistent, contextually relevant responses to a wide range of natural language input. Figure \ref{fig:ChatGPT_Diagram} shows the ChatGPT training process \citep{openai2022chatgpt}.

\section{Designing our Exploratory Study}\label{sec:study}


For this study, we focus on using SPC in three areas relevant to SPC: practice, learning, and research. We evaluate ChatGPT's ability (1) to provide code for a specified task; (2) to explain basic concepts; and (3) to create knowledge related to each of the three areas. 

\begin{figure}
    \centering
    \includegraphics[width=0.95\textwidth, frame]{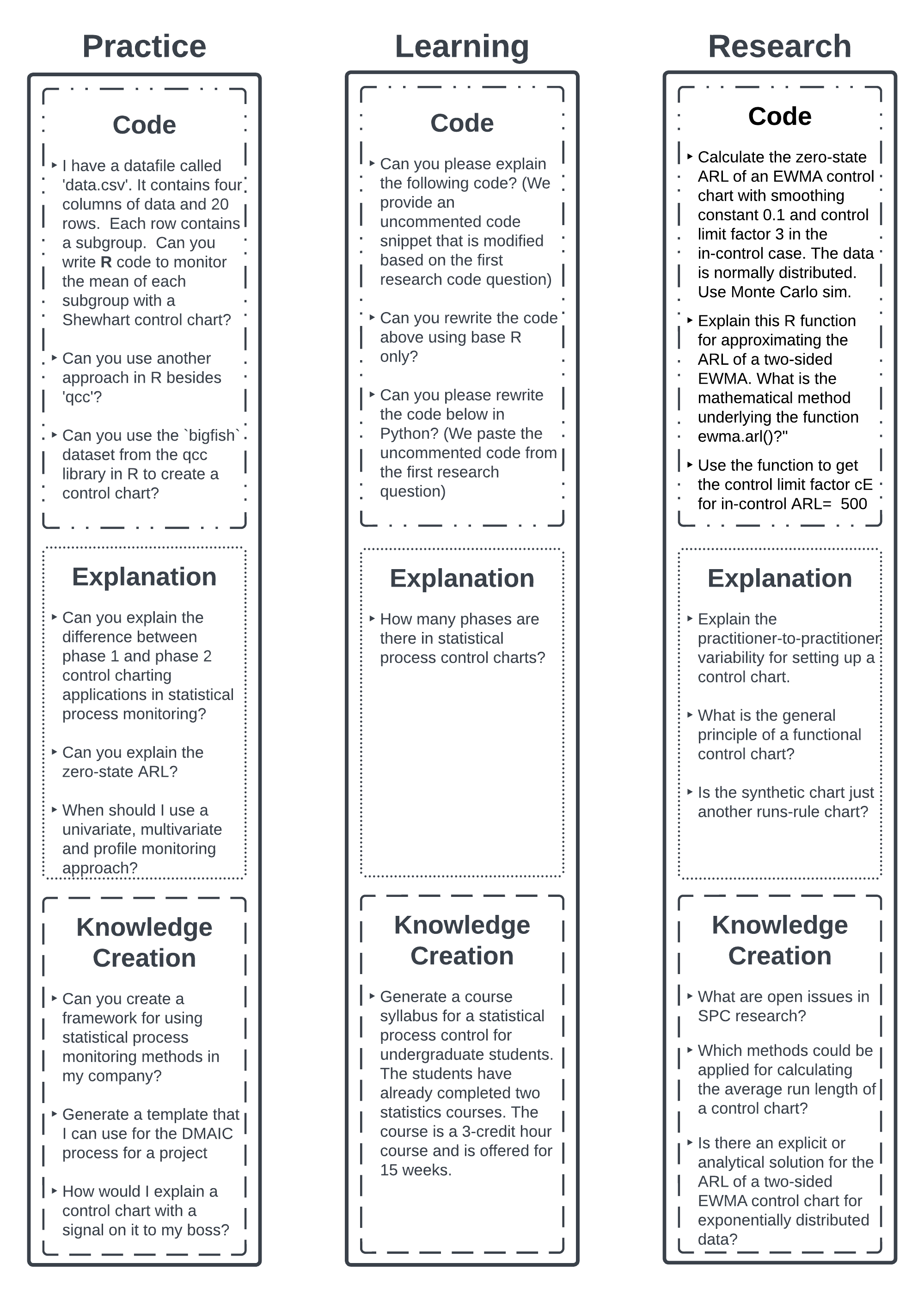}
    \caption{An overview of our study design.}
    \label{fig:study_design}
\end{figure}

To demonstrate ChatGPT's ability to generate code, we evaluated several prompts related to creating control charts for practice, explaining existing code, translating code between languages for learning, and developing simulation code for research. To investigate ChatGPT's ability to provide correct and meaningful explanations of key concepts, we evaluated prompts related to defining and explaining SPC terms in practice, learning, and research. Finally, to assess the ChatGPT's ability to create new knowledge assets, we prompted it to create a framework for SPC, a syllabus for an SPC course, and to determine open research issues in SPC. The exact prompts used appear in Figure \ref{fig:study_design} and Sections \ref{sec:practice}, \ref{sec:learning}, and \ref{sec:research}.

Our team of experienced educators, authors, and statisticians evaluated the responses. Currently, ChatGPT can provide but not execute code. Therefore, our team copied and pasted the code into an appropriate environment (\texttt{R} or \texttt{Python}), attempted to run the code, and evaluated the correctness of the output. In cases where the code would not execute, our team corrected the code, explained the errors, and assessed the correctness of the output. We compared the ChatGPT responses to current textbooks and well-respected literature on the topics to evaluate the prompts related to explaining ideas and concepts. Similarly, we assessed the responses for knowledge creation and compared them to existing knowledge and similar assets for each domain area (practice, learning, and research). 

Our methods are qualitative and are naturally colored by our own biases and experience in the area of SPC. A characteristic of generative AI models like ChatGPT is that responses to specific prompts are not repeated (which is why we have included many screenshots for documentation). 

\section{On the Use of ChatGPT by SPC Practitioners}\label{sec:practice}

\label{practitioners}

We explore how ChatGPT can be used to augment coding, explain concepts, and create knowledge assets for practitioners. In the subsections below, we provide the results generated by our ChatGPT prompt, followed by our interpretation of ``what worked'' and ``what failed''.  

\subsection{Code}
\begin{figure}[htb!]
    \centering
    \includegraphics[width=\textwidth, trim = {0 2.25in 0 0.05in}, frame, clip]{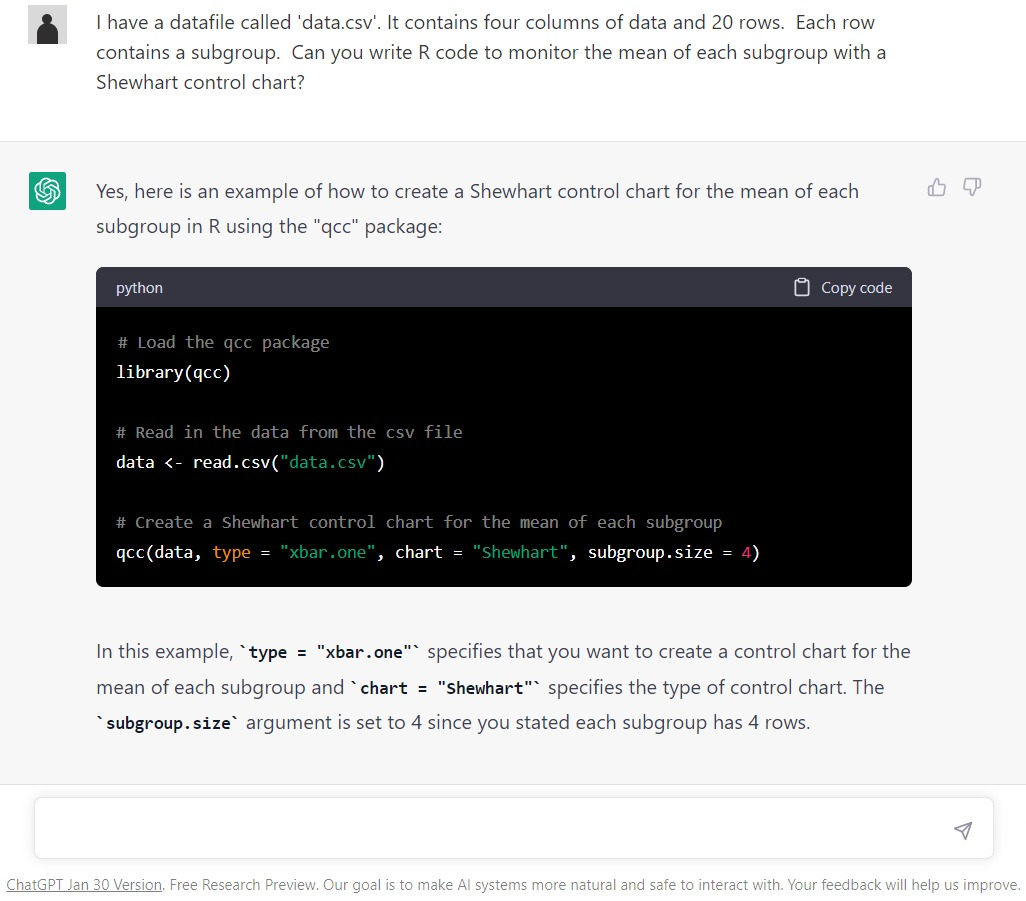}
    \caption{A screenshot of ChatGPT's response to reading the `data.csv' file and writing an \texttt{R} code to monitor the mean of each subgroup with an Xbar Chart.
    }\label{fig:practiceprompt1}
\end{figure}

First, we asked ChatGPT to write \texttt{R} code that can be used to create a Shewhart $\Bar{X}$ chart based on a `data.csv' file containing 20 subgroups and four observations per subgroup. Figure \ref{fig:practiceprompt1} depicts our exact prompt and the corresponding code. From the output, ChatGPT incorrectly named the code chunk \texttt{Python} even though the code was output in \texttt{R}. ChatGPT selected a suitable \texttt{R} library, \texttt{qcc}, which can be used to construct a $\Bar{X}$ chart. The code assumes that the package is already installed on the user's machine or that the user has a cursory knowledge of \texttt{R} to know that the \texttt{library()} function call would only work if the package has been installed in the past. Otherwise, the first line of code would return the following: ``\texttt{Error in library(qcc): there is no package called `qcc'}.'' Then, the code assumes that the `data.csv' is in the user's working directory. In our estimation, both assumptions are reasonable given that the prompt specifically asked for \texttt{R} code. 
However, running the code results in the following error: ``\texttt{Error in if (chart.all) \{ : argument is not interpretable as logical}.''       

\begin{figure}[htb!]
    \centering
    \begin{framed}
    \begin{subfigure}[b]{\textwidth}
        \centering
        \includegraphics[width=0.95\textwidth]{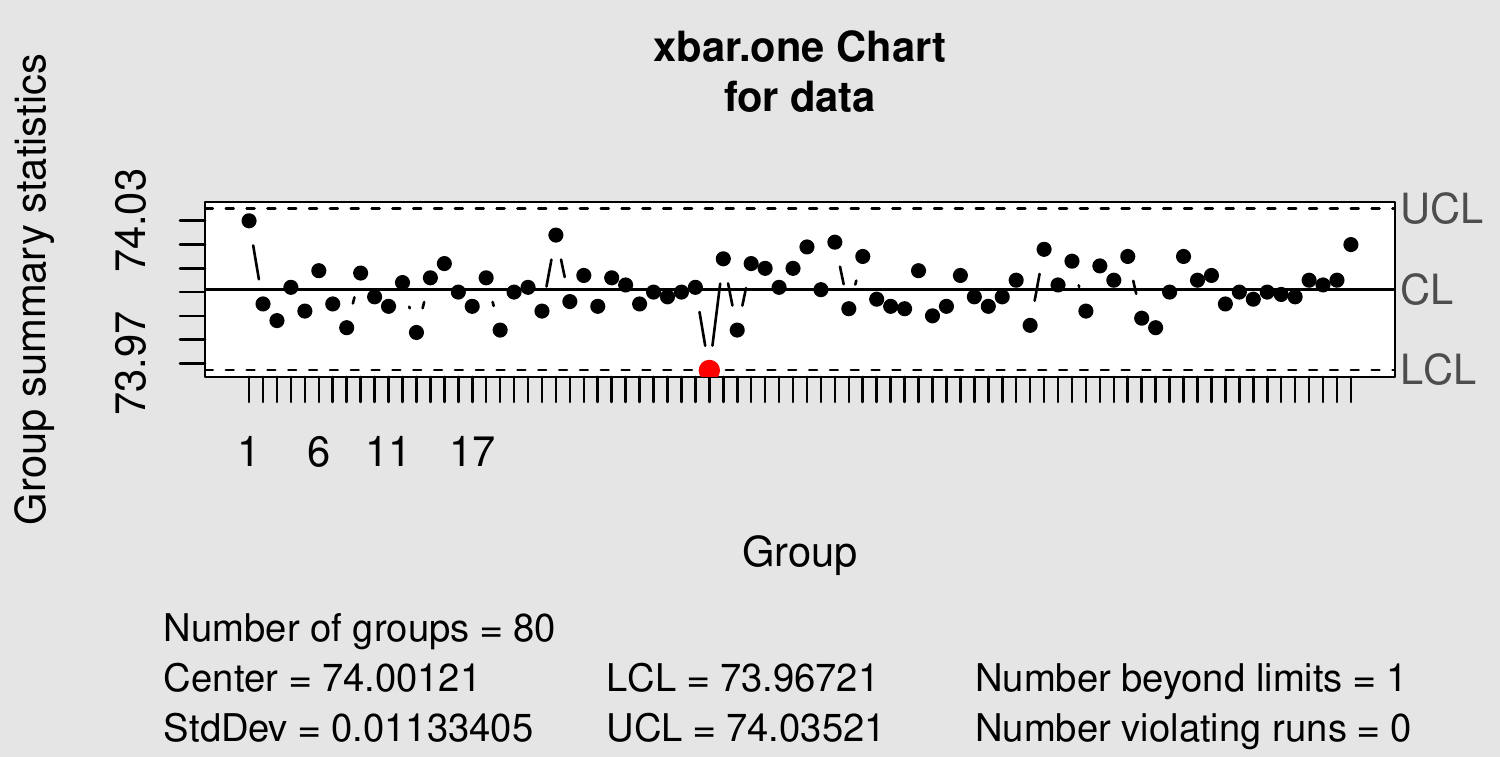}
        \caption{The incorrect chart, no subgrouping of the data, produced after eliminating the error message obtained when running the third line of the ChatGPT code.}
        \label{fig:incorrect_xbar}
    \end{subfigure}

    \vspace{\baselineskip}
    
   \begin{subfigure}[b]{\textwidth}
        \centering
        \includegraphics[width=0.95\textwidth]{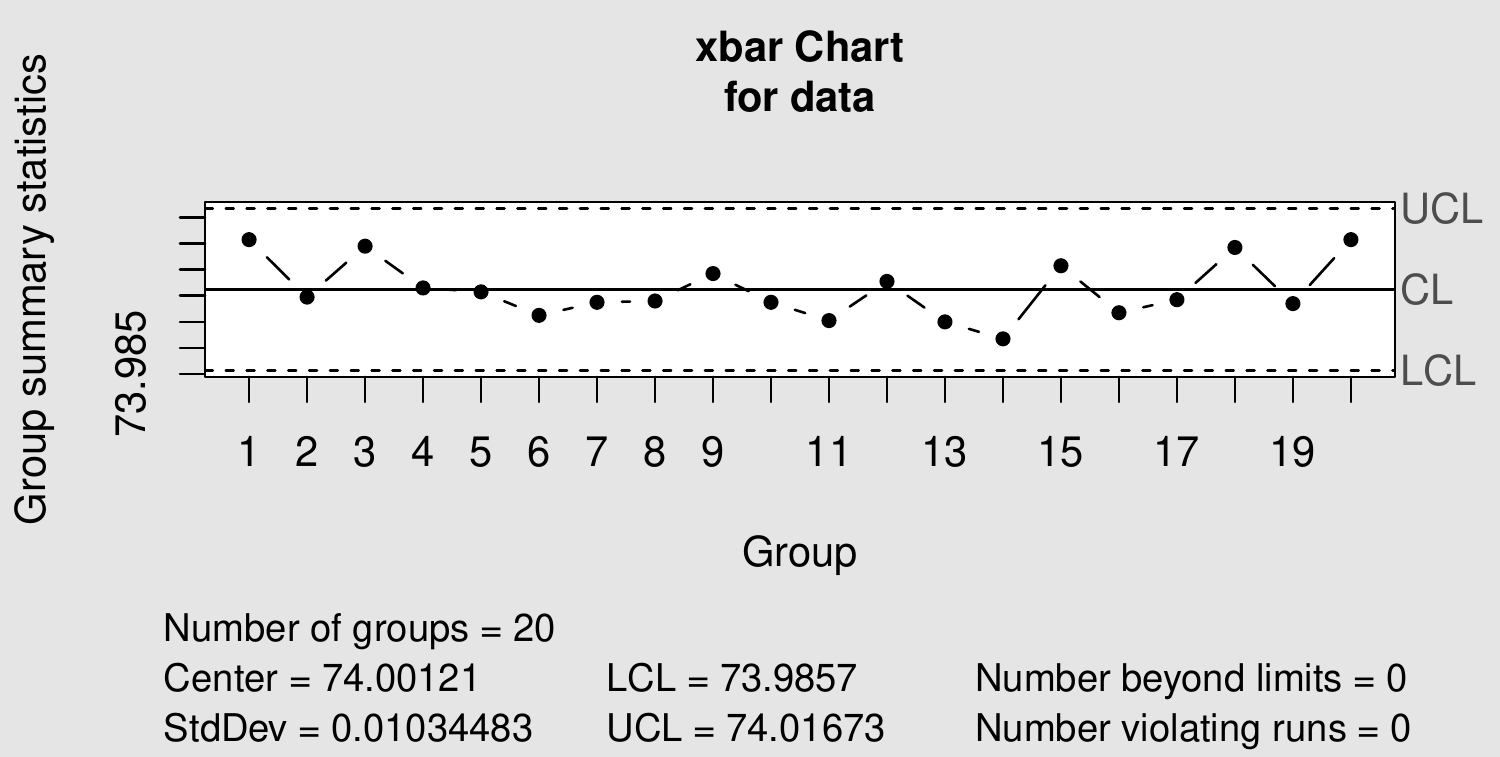}
        \caption{The correct $\Bar{X}$ chart, with subgrouping of the data, produced after changing assigning the type argument to `xbar' instead of `xbar.one`.}
        \label{fig:correct_xbar}
    \end{subfigure}
    \end{framed}

    \vspace{-\baselineskip}
    
    \caption{A comparison of the two produced charts, with slight modification to the arguments in the \texttt{qcc()} function used by ChatGPT.}
    \label{fig:xbar}
\end{figure}

We attempted to understand why the code did not run. This process involved two steps. First, we tried eliminating the error and assigned the \texttt{chart} argument to a logical (e.g., \texttt{chart = TRUE}) based on the printed error message. This eliminated the error message but resulted in an incorrect chart (an individual chart instead of the desired $\Bar{X}$ chart, see Figure \ref{fig:incorrect_xbar}). Second, we produced a correct chart, depicted in Figure \ref{fig:correct_xbar}, by changing the \texttt{type} argument's value from `xbar.one' to `xbar'. We do not anticipate that the obtained error results from different package versions, as we checked earlier package releases from \citet{scrucca2004qcc,scrucca2017tour}.   

ChatGPT did not correctly assign arguments to the \texttt{qcc()} function: the values assigned to the \texttt{type} and \texttt{chart} arguments were incorrect. Moreover, the \texttt{subgroup.size} argument itself seems erroneous based on the source code for versions 1.0, 2.0, and 2.7 of the package. 
The argument's correct name should have been \texttt{sizes}. It is also not needed based on the argument's definition: ``{a value or a vector of values specifying the sample sizes associated with each group. For continuous data provided as a data frame or matrix, the sample sizes are obtained by counting the non-NA elements of each row. For `p', `np', and `u' charts, the argument sizes is required.}'' 

\begin{figure}[htb!]
    \centering
    \includegraphics[width=0.95\textwidth, frame]{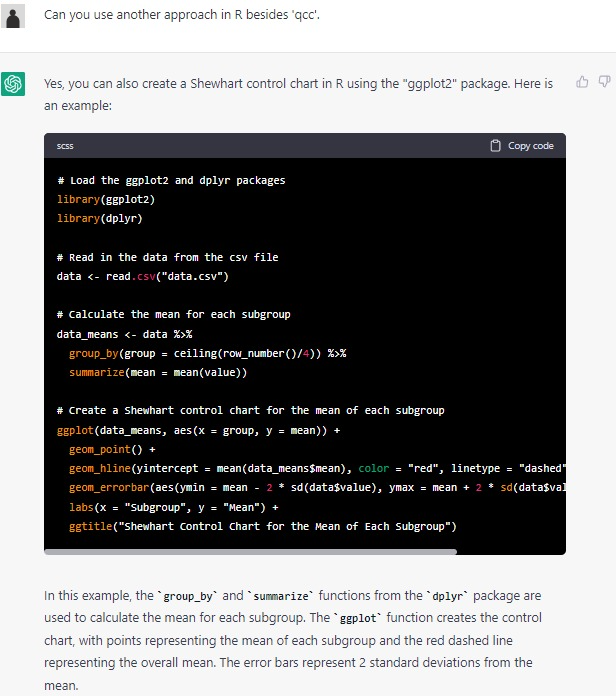}
    \caption{A screenshot of ChatGPT's response for our second practitioner-focused code prompt. 
    }\label{fig:practiceprompt2}
\end{figure}
\begin{figure}[htb!]
    \centering
    \begin{framed}
    \begin{subfigure}[b]{\textwidth}
        \begin{lstlisting}[basicstyle =\footnotesize, 
        language = R,
        backgroundcolor = \color{light-gray},
        keywords={group\_by, row\_number, mutate, ggplot, aes, geom\_point, geom\_line, geom_hline, mean, max, min, labs, ggtitle, c, |>, -, +, read.csv, library, ungroup, read, csv},
        keywordstyle=\bfseries,
        otherkeywords={!,!=,~,$,*,\&,\%/\%,\%*\%,\%\%,<-,<<-,/, +, -},
        alsoother={.$},
        showstringspaces=false,
        breaklines=true]
# Load the ggplot2 and dplyr packages
library(ggplot2)
library(dplyr)

# Read in the data from the csv file
data <- read.csv("data.csv")

# Calculate the mean for each subgroup
data_means <- data |>
  group_by(group = row_number()) |> 
  mutate(
    avg = mean( c(obs_1, obs_2, obs_3, obs_4) ),
    R = max(c(obs_1, obs_2, obs_3, obs_4)) - min(c(obs_1, obs_2, obs_3, obs_4))
  ) |> 
  ungroup()

# We do not recommend the use of ranges to estimate the standard deviation
# but we have used this approach here to reproduce the output from the qcc plot
# 0.729 is the control limit factor for the x-bar chart when subgroup size = 4 
ggplot(data_means, aes(x = group, y = avg)) +
  geom_point(size = 2) +
  geom_line(size = 1) +
  geom_hline(yintercept = mean(data_means$avg), linetype = "solid") +
  geom_hline(yintercept = mean(data_means$avg) + 0.729*mean(data_means$R), linetype = "dashed") +
  geom_hline(yintercept = mean(data_means$avg) - 0.729*mean(data_means$R), linetype = "dashed") +
  labs(x = "Subgroup", y = "Mean") +
  ggtitle("xbar Chart for data")
        \end{lstlisting}
        \caption{One possible approach, which capitalizes on the \texttt{ggplot2} and \texttt{dplyr} packages, to generate a \textit{qcc}-equivalent $\Bar{X}$ chart while fixing the errors from the outputted ChatGPT code.}
        \label{fig:code_listing}
    \end{subfigure}

    \vspace{\baselineskip}
    
   \begin{subfigure}[b]{\textwidth}
        \centering
        \includegraphics[width=0.95\textwidth]{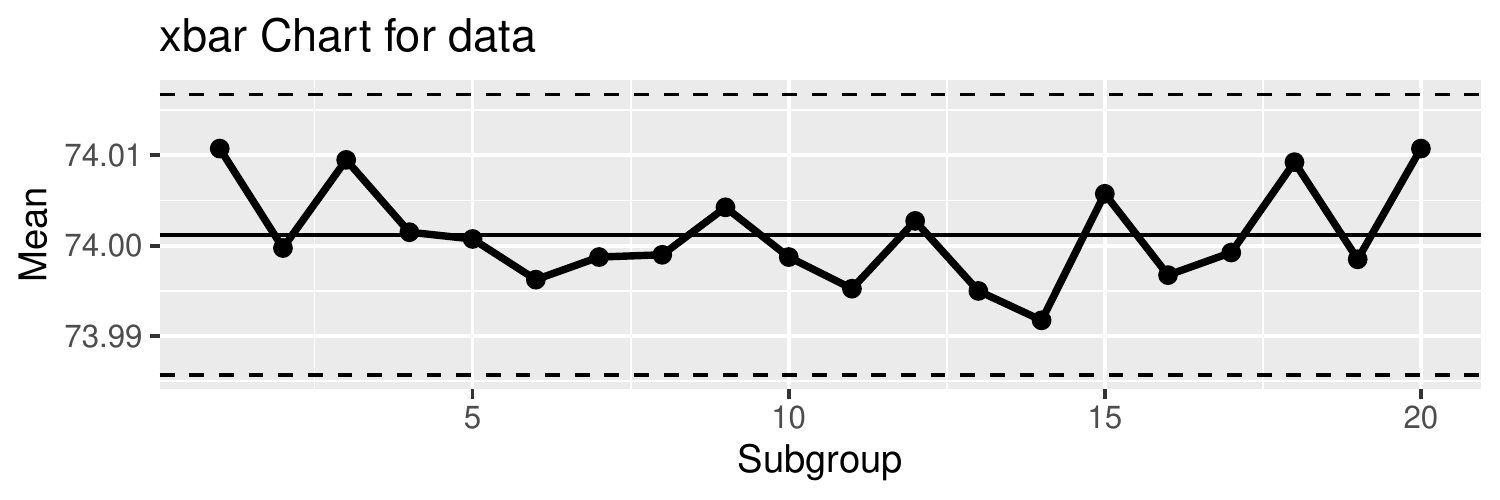}
        \caption{The correct $\Bar{X}$ chart, with subgrouping of the data, produced after we significantly changed the ChatGPT code in Practice Prompt \# 2.}
        \label{fig:correct_gg_xbar}
    \end{subfigure}
    \end{framed}

    \vspace{-\baselineskip}
    
    \caption{Our \texttt{dplyr} and \texttt{ggplot2}-based code and the resulting $\Bar{X}$ chart, which is equivalent to that produced by the \textit{qcc} package. Our `data.csv' is a sample of the \texttt{pistonrings} dataset (we only keep the first 4 out of 5 observations for subsamples $1-20$), and can be downloaded from the link provided in the \textit{Online Materials} section.}
    \label{fig:ggcode_and_plot}
\end{figure}

As a follow-up, we have examined whether ChatGPT can provide an alternate method for constructing the $\Bar{X}$ chart in \texttt{R}. The results are shown in Figure \ref{fig:practiceprompt2}. Similar to the first prompt, the code starts by calling \texttt{library()} to load the required packages. The comments in the code highlight the steps needed to create a Shewhart chart using the \textit{tidyverse} approach. However, similar to the previous code prompt, the resulting code is erroneous and will not produce correct results. Several issues were noted upon debugging the generated \texttt{R} code: 
\begin{enumerate}[label=(\alph*), nosep]
    \item the attempted \texttt{group()} operation is incorrect since it assumes that every observation within a subgroup is in a separate row. 
    \item the object value in the subsequent line was never defined, which would result in an error due to \texttt{``object `value' not found''}; and
    \item neglecting the two previous errors, the use of the \texttt{geom\_errorbar} is also incorrect in this instance.
\end{enumerate}
If we were to grade ChatGPT on consistency, it is unclear why two-sigma limits were used with \texttt{geom\_errorbar} when the default 3-sigma limits were used in the `qcc' function.  
Figure \ref{fig:ggcode_and_plot} presents our modified code and the resulting \texttt{ggplot}, with limits matching (with rounding error) those produced by the \texttt{qcc} package.

\begin{figure}[htb!]
    \centering
    \begin{framed}
    \begin{subfigure}[b]{\textwidth}
        \centering
        \includegraphics[width=\textwidth, trim = {0 4.5cm 0 0}, clip]{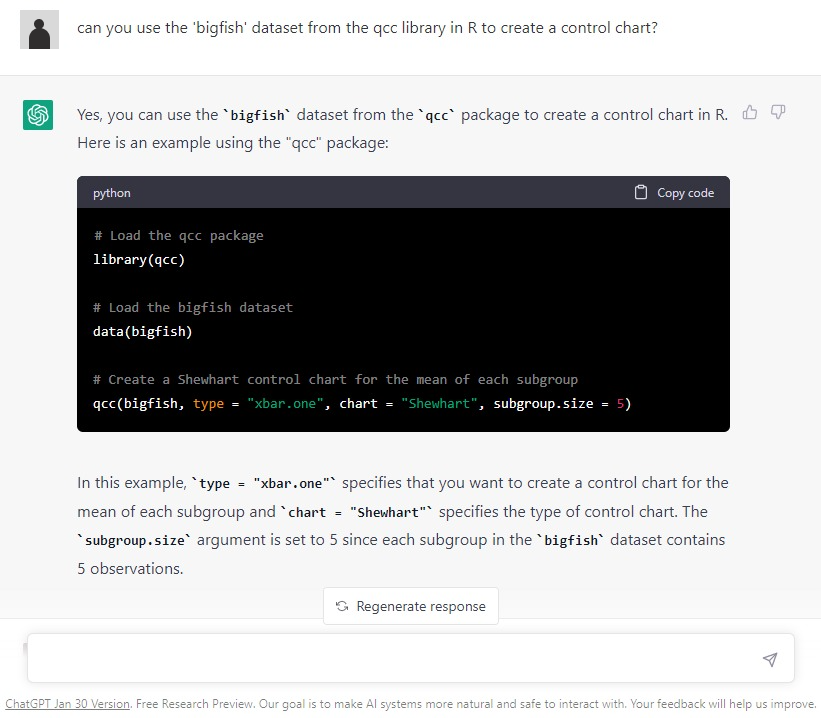}
        \caption{ChatGPT's `explanation' of how the \texttt{qcc} package can be used to chart the `bigfish' dataset.}
        \label{fig:bigfish01}
    \end{subfigure}

    \vspace{\baselineskip}
    
   \begin{subfigure}[b]{\textwidth}
        \centering
        \includegraphics[width=\textwidth, trim = {0 7.5cm 0 0}, clip]{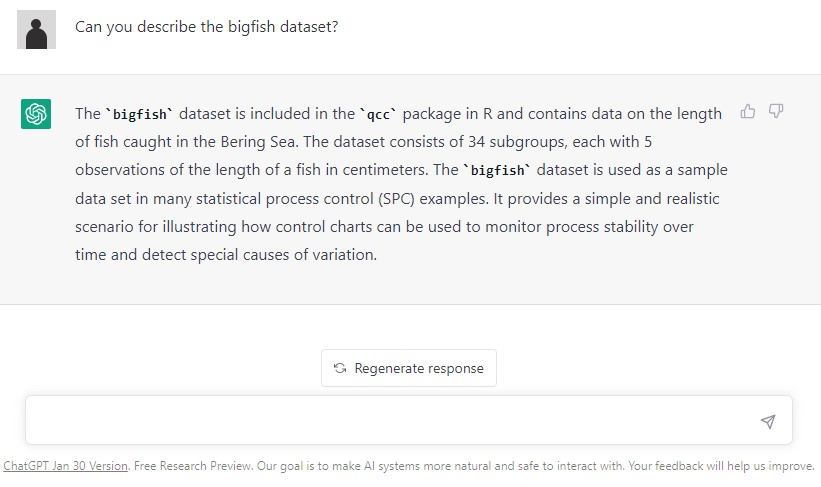}
        \caption{ChatGPT makes up a somewhat compelling yet fake description of the `bigfish' dataset.}
        \label{fig:bigfish02}
    \end{subfigure}
    \end{framed}

    \vspace{-\baselineskip}
    
    \caption{ChatGPT is susceptible to simple adversarial attacks: A `bigfish' story. 
    }
    \label{fig:bigfish}
\end{figure}

In the third and final practical code evaluation, we attempt to see if ChatGPT can detect erroneous requests. Given its apparent familiarity with the \texttt{qcc} package, we asked it to ``can you use the ‘bigfish' dataset from the qcc library in \texttt{R} to create a control chart?'' Note that we made up this dataset name. Furthermore, we ensured that it does not exist in the package via the \texttt{R} command \texttt{data(package = `qcc')}, which showed all the datasets within the \texttt{qcc} package (i.e., `bigfish' was not in the printed list). In its initial output, ChatGPT utilized the exact solution approach presented in Practice Prompt \#1. It modified that code (see Figure \ref{fig:bigfish01}) to load the non-existent dataset via the \texttt{data(bigfish)} function (which would have worked if the `bigfish' dataset was indeed in the \texttt{qcc} package). In our estimation, ChatGPT's use of \texttt{data(bigfish)} can be attributed to the use of reinforcement learning from human feedback (RLHF) to fine-tune the underlying GPT-3.5 model. While the RHLF approach was pioneered by OpenAI so that the LLMs are much better at following human intentions \citep{ouyang2022training}, our intentionally misleading question can be described as an adversarial attack on the model. 
However, the third line of code and the subsequent explanation were troubling since ChatGPT decided that the ``\texttt{subgroup.size} argument is set to 5 since each subgroup in the `bigfish' dataset contains 5 observations.'' So we asked it a follow-up question that was not pre-planned. Our question and the ChatGPT explanation are depicted in Figure \ref{fig:bigfish02}. The ``making up'' of a seemingly accurate description based on a non-existent dataset within the package is also troubling since: (a) we could not find a ``bigfish'' dataset with such a description on the internet, and (b) the made-up description is believable (i.e., it clearly states why this dataset is included in the package, with a reasonable description of the number of subgroups and samples).  

\subsection{Explanation}
\label{sec:practice_explanation}

In this subsection, we examine ChatGPT's ability to explain three concepts: (a) the difference between Phase 1 and Phase 2 control charting applications, (b) the zero-state ARL, and (c) when one should use a univariate, multivariate, and profile monitoring approach. The text-based responses for ChatGPT are depicted in Figure \ref{fig:practice_explain}.

\begin{figure}[htb!]
    \centering
    \begin{framed}
    \begin{subfigure}{\textwidth}
    \centering \includegraphics[width=0.92\textwidth]{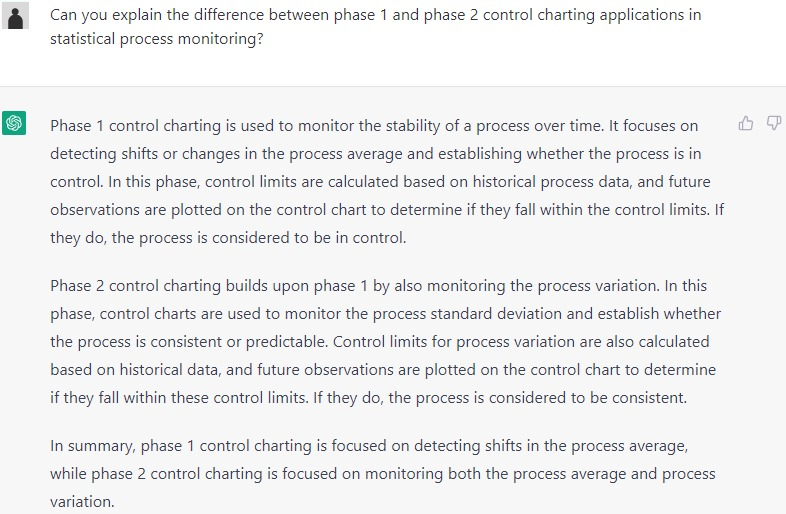}
    \subcaption{ChatGPT's explanation of Phase 1 and Phase 2 control charting applications.}
    \label{fig:phase1_2}
    \end{subfigure}

    \vspace{\baselineskip}
    
    \begin{subfigure}{\textwidth}
    \centering
    \includegraphics[width=0.92\textwidth]{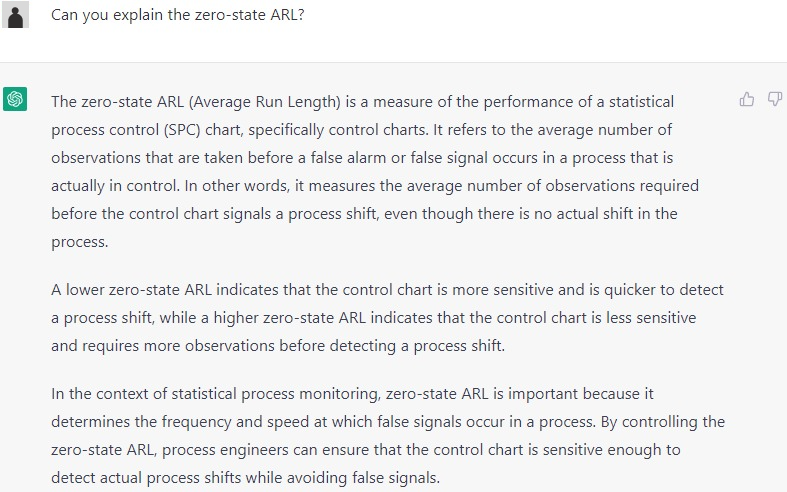}
        \subcaption{ChatGPT's explanation of the zero-state ARL.}
        \label{fig:zero_state_ARL}
    \end{subfigure}
    \end{framed}
\end{figure}%
\begin{figure}[htb!]\ContinuedFloat
    \centering
    \begin{framed}
    \begin{subfigure}{\textwidth}
    \includegraphics[width=0.92\textwidth]{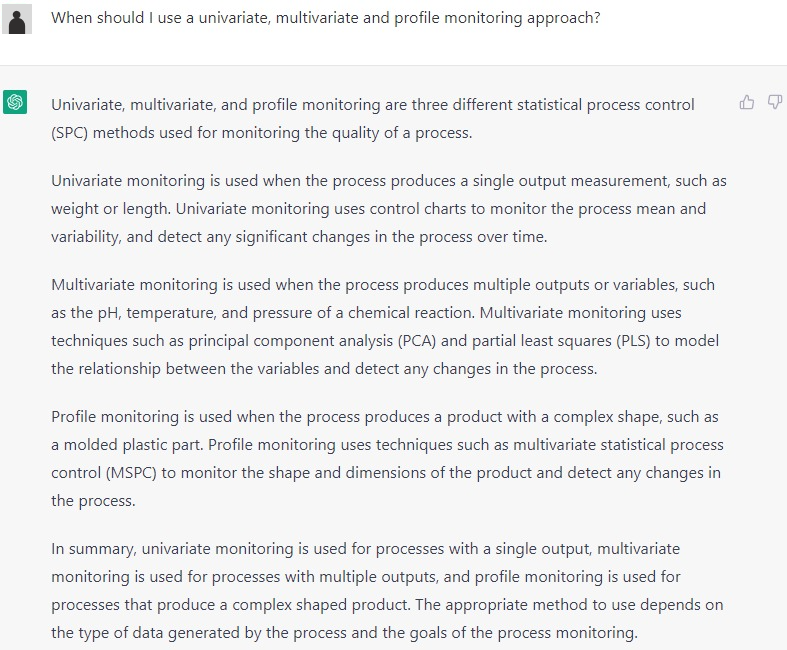}
        \subcaption{ChatGPT's explanation of when to use univariate, multivariate, and profile monitoring approaches.}
        \label{fig:choice_method}
    \end{subfigure}
    \end{framed}

    \vspace{-\baselineskip}
    
    \caption{ChatGPT's response to our three explanatory practitioner prompts. 
    }
    \label{fig:practice_explain}
\end{figure}

The first prompt asked ChatGPT to ``explain the difference between phase 1 and phase 2 control charting applications in statistical process monitoring.'' The response, shown in Figure \ref{fig:practice_explain}, ChatGPT incorrectly stated that Phase 1 control charting is used to monitor the stability of the process over time.  This is incorrect and more closely aligns with accepted definitions of Phase 2 of control charting.  ChatGPT also responded that Phase 1 is focused on detecting changes in the process average; however, Phase 1 is not limited to evaluating the process average.  Although it correctly stated that Phase 1 control charting is used to establish whether the process is in control and calculate control limits, it incorrectly stated that future observations are monitored over time in Phase 1.  It is widely accepted that Phase 1 involves retrospective analysis of a process to evaluate process variability, checking for outliers, defining the in-control state of the process, developing a process model, estimating model parameters, and establishing a control chart for prospective monitoring \citep{jones2014overview}.  

When defining Phase 2, ChatGPT begins incorrectly, stating that Phase 2 builds upon Phase 1 by monitoring the process variation.  The definition of Phase 2 given by ChatGPT is similar to that of Phase 1 but focuses solely on monitoring variation.  It is widely accepted that in Phase 2 of control charting, one uses the process model developed in Phase 1 to monitor prospectively for departures from the model.  Phase 2 analysis is not limited to monitoring the process variability as implied by ChatGPT, but can be applied to any important process parameter identified in Phase 1.  The final summary of the ChatGPT response states that Phase 1 control charting focuses on shifts in the average, while Phase 2 focuses on monitoring both the average and the variation.  This is incorrect.  For more details on Phase 1 and 2 of process monitoring see, e.g., \citep{montgomery2020introduction,chakraborti2008phase,jones2014overview,zwetsloot2023univ}.

The second prompt asked ChatGPT to ``explain the zero-state ARL.''  The commonly accepted definition of the zero-state ARL is the average number of plotted statistics counted from the start of monitoring before a control chart signals \citep{Cros:1986, Knot:2021c}. 
This is contrasted with the steady-state ARL, which measures the average number of plotted statistics until a signal is observed with counting beginning when the process is in steady state. A process is defined to be in steady-state if ``the process has been in control long enough for the effect of the starting value [of the chart statistic] to become negligible'' \citep{lucas1990exponentially}. 

When given the prompt to explain the zero-state ARL, ChatGPT provided a partially correct answer but either ignored or misinterpreted the qualifier ``zero-state''.  The basic definition given by ChatGPT is a mostly correct definition of the in-control (not zero-state) ARL as the ``average number of observations that are taken before a false alarm or a false signal occurs in a process that is actually in control.''  The accepted definition of the ARL is the average number of \textit{plotted statistics} observed before a signal is observed.  The use of the term \textit{plotted statistic} in place of \textit{observations that are taken} is more general and can be applied to subgrouped data.  Adding to this definition the average number of plotted statistics observed before a \textit{false} signal is observed provides a correct definition of the in-control ARL.  ChatGPT then restates the definition: ``it measures the average number of observations required before the control chart signals a process shift, even though there is no actual shift in the process.''  Once again, this is a near-correct description of the in-control ARL, but there is no mention of when counting begins to indicate zero-state. Thus, we can likely conclude that ChatGPT misinterpreted the \textit{zero-state} qualifier for the ARL as \textit{in-control}.  

In the next statement describing the ARL, ChatGPT correctly stated that a lower ARL indicates a chart that is quicker to detect a process change, and a chart with a higher ARL requires more observations to detect a change. In addition, ChatGPT correctly contextualized the ARL as a tool for process engineers to understand the sensitivity of a chart to changes, the frequency of signals, and the usefulness of balancing the detection of changes with avoiding false signals.  

The third prompt provided to ChatGPT asked ``When should I use univariate, multivariate, and  profile monitoring?''  ChatGPT correctly noted that univariate monitoring is used to monitor a single process measure over time.  The answer given, however, is specific to variables data, noting that one monitors the mean and variance with univariate control charts.  A more general answer would include reference to, e.g., proportions, counts, and location measure other than the mean.  

The advice for multivariate monitoring is also correct, noting that it is used for monitoring multiple process variables.  Using a simple example of monitoring the pH, temperature, and pressure of a chemical reaction, ChatGPT suggests that one would use methods such as principal components analysis (PCA) or partial least squares (PLS) for multivariate monitoring.  While these methods are useful for monitoring high-dimensional processes, the most widely used multivariate control chart, Hotelling's $T^2$ chart was not mentioned.  

ChatGPT also correctly identified the context for profile monitoring to monitor a shape such as that of a molded plastic part.  It qualified this, however, to be a \textit{complex shape}.  In actuality, profile monitoring can be used for both simple and complex shapes, and monitoring linear profiles is common in practice \citep{kim2003monitoring}.  ChatGPT states that multivariate statistical process control is used to monitor the shape and dimensions of the profile.  This is a correct, but very general statement.  For reviews of the research on profile monitoring, see, \citep{saghaei2013statistical,maleki2018overview,woodall2007current}


\subsection{Knowledge Creation}

In this section, we ask ChatGPT to provide descriptions that can be used in reporting SPC concepts in practice. We asked it the following questions:
\begin{enumerate}[label = (\alph*), nosep]
    \item Can you create a framework for using statistical process monitoring methods in my company?
    \item  Generate a template that I can use for the DMAIC process for a project.
    \item How would I explain a control chart with a signal on it to my boss?
\end{enumerate}
The generated responses were reasonable, similar to our discussion in Section \ref{sec:practice_explanation}.   

For question (a), ChatGPT presented a framework made of the following seven steps: (i) define the process, (ii) collect data, (iii) choose a control chart, (iv) establish the control limits, (v) monitor the process, (vi) detect and respond to out-of-control conditions, and (vii) continuously evaluate and improve. For each step, it provided a 1-2 sentence explanation and concluded that the framework might need to be modified based on the company's requirements. Similarly, for question (b), ChatGPT identified 2-3 tasks that should be completed within each step of the DMAIC process. Its conclusion followed the previous question, stating that this is a general framework that may need to be tweaked based on the application domain. Note that we do not provide snapshots of its responses here for conciseness. The interested reader should refer to our \textit{Supplementary Materials} for the recordings of the ChatGPT answer.

\begin{figure}[htb!]
    \centering
    \includegraphics[width=0.95\textwidth, frame]{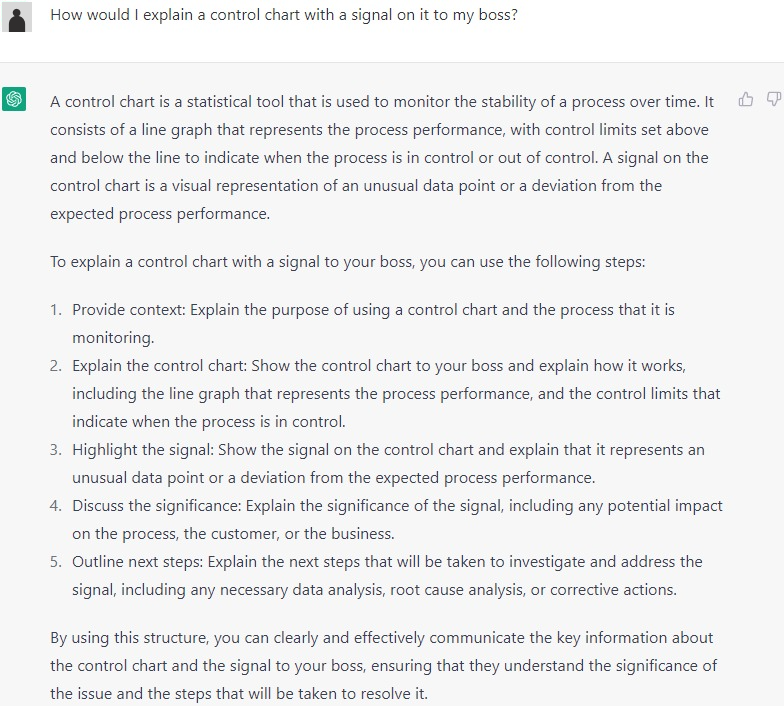}
    \caption{ChatGPT's answer on how to explain a control chart signal to your boss.}
    \label{fig:chart_signal}
\end{figure}

ChatGPT provided a compelling answer to the last question (see Figure \ref{fig:chart_signal}). While we did not explicitly ask for steps, the AI likely learned from the two previous questions and provided an excellent breakdown of the question. Its introductory paragraph provided working definitions for control charts and signals, and then it highlighted five steps that should be followed when explaining a signal to one's boss. It wrapped up by emphasizing the need to communicate the principles behind a control chart and signal effectively while ensuring the significance and the steps taken to resolve the signal.

While we appreciate the points made by ChatGPT in its answer, we note that the output lacks sufficient statistical detail. This can partially be attributed to the generalized nature of our input prompt/question. However, it also highlights the differences between a chat with an AI versus a statistical consultant. In the AI case, the questions are only asked by the human interacting with ChatGPT. However, in a human-to-human interaction, the statistical consultant would have likely asked several questions to tailor their response to the audience. These questions would have likely included the following: 
\begin{itemize}[nosep]
    \item What are the boss's title and training background? 
    \item Was this signal observed in Phase I or Phase II?
    \item What type of control chart was used?
    \item How were the control limits determined?
    \item What is the application domain (e.g., in public health surveillance, the control chart does not typically reset after a Phase II signal)? 
    \item Can the process be stopped until a thorough investigation of the process occurs?
\end{itemize}
From these answers, the consultant would be able to generate a more precise response. When interacting with an AI chatbot, the onus is on the human to provide this context upfront or ask the AI to tweak its answer by supplementing its responses with additional information.

\section{Learning SPC Fundamentals with ChatGPT}\label{sec:learning}

\subsection{Code}
To examine whether ChatGPT can accelerate learning in the context of coding, we have designed three coding tasks. These three coding tasks relate to understanding code, providing an alternative approach to solving a problem with code, and translating code from one language to another. These questions and ChatGPT's corresponding response are detailed in the following paragraphs.

\begin{figure}[htb!]
    \centering
    \includegraphics[width=0.92\textwidth, frame, trim ={0 0.2cm 0 0.4cm}, clip]{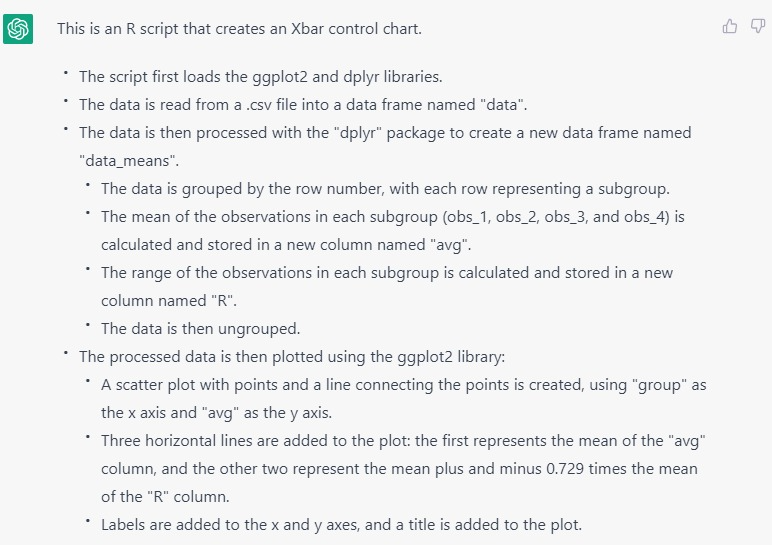}

    \vspace{-0.5\baselineskip}
    
    \caption{ChatGPT's explanation of an uncommented version of the code shown in Figure \ref{fig:code_listing}. We provided the code in a new chat window so that ChatGPT would have no context for our previous conversations. 
    }
    \label{fig:learning_explain_code}
\end{figure}

In our first question, we provided ChatGPT with an uncommented version of our code prompt from Figure \ref{fig:code_listing} and asked to explain the code. This question was asked in a new chat window so that no context from our previous discussions is given. ChatGPT provided a comprehensive explanation of our code, as shown in Figure \ref{fig:learning_explain_code}. It clearly captured the different steps and provided an explanation of the different functions. It even recognized that this code is used for the creation of an $\Bar{X}$ control chart (possibly based on the inputs from the \texttt{ggtitle()} function). One missing aspect from its explanation was to highlight what the three horizontal lines stand for, i.e., that the \texttt{geom\_hline()} was used to draw the center line (with the solid line) and the lower control limit (LCL)/upper control limit (UCL) with the dashed lines. Overall, we are pleasantly surprised by its explanation. 

\begin{figure}[htb!]
    \centering
    \begin{framed}
    \begin{subfigure}[b]{\textwidth}
        \begin{lstlisting}[basicstyle =\footnotesize, 
        language = R,
        backgroundcolor = \color{light-gray},
        escapeinside={(*@}{@*)},
        keywords={group\_by, row\_number, mutate, ggplot, aes, geom\_point, geom\_line, geom_hline, mean, max, min, labs, ggtitle, c, |>, -, +, read.csv, library, ungroup, read, csv, plot, abline, title, apply, rowMeans, function, row.names, data.frame},
        keywordstyle=\bfseries,
        otherkeywords={!,!=,~,$,*,\&,\%/\%,\%*\%,\%\%,<-,<<-,/, +, -},
        alsoother={.$},
        showstringspaces=false,
        breaklines=true]
data <- read.csv("data.csv")

(*@  \hldgray{group <- \textbf{row.names}(data)  \# Changed from ChatGPT's incorrect:  group <- \textbf{row\_number}(data)}  @*)

data_means <- (*@ \textbf{data.frame}@*)(
    group = group,
    avg = rowMeans(data[, c("obs_1", "obs_2", "obs_3", "obs_4")]),
    R = apply(data[, c("obs_1", "obs_2", "obs_3", "obs_4")], 1,
                function(x) max(x) - min(x) )
)

plot(data_means$group, data_means$avg, type = "b", pch = 19, cex = 2, 
    xlab = "Subgroup", ylab = "Mean")
abline(h = mean(data_means$avg), lty = 1)
abline(h = mean(data_means$avg) + 0.729 * mean(data_means$R), lty = 2)
abline(h = mean(data_means$avg) - 0.729 * mean(data_means$R), lty = 2)
title(main = "xbar Chart for data")
        \end{lstlisting}
        \caption{The base R code generated by ChatGPT in lieu of Figure \ref{fig:code_listing}. The darker second row highlights a minor change to ChatGPT's code where we used the \texttt{base R} function \texttt{row.names} instead of the incorrect \texttt{dplyr} function \texttt{row\_number}.}
        \label{fig:baseR_listing}
    \end{subfigure}

    \vspace{\baselineskip}
    
   \begin{subfigure}[b]{\textwidth}
        \centering
        \includegraphics[width=0.95\textwidth, trim = {0 0.2in 0 0.25in}, clip, frame]{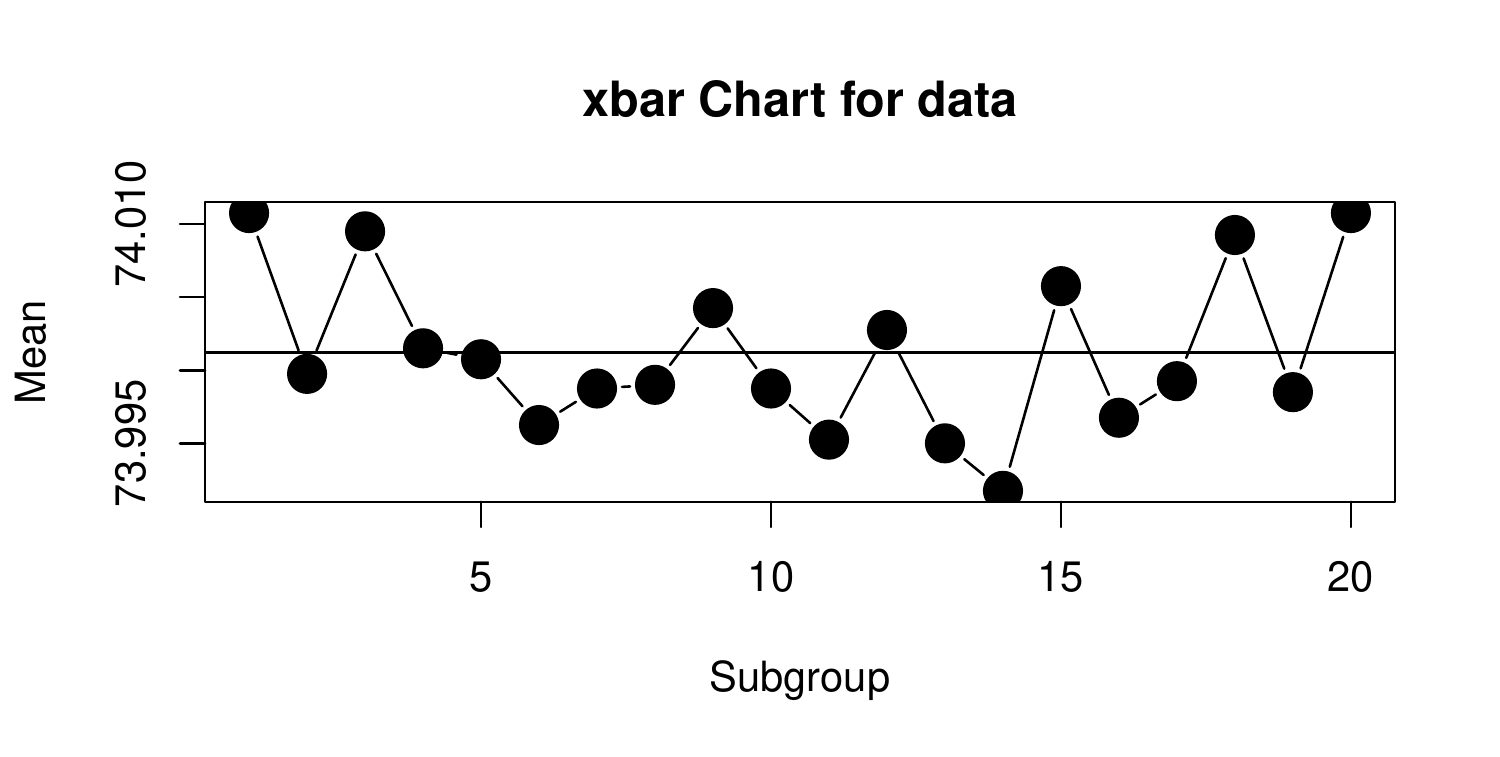}
        \caption{The correct $\Bar{X}$ chart produced with the minor modification of the ChatGPT code. Note that the UCL and LCL overlap with the existing boundaries for the chart. Hence, to show them, the reader can set the \texttt{ylim} argument in the plot to an appropriate value (e.g., \texttt{ylim = c(73.98, 74.02)}.)}
        \label{fig:baseR_plot}
    \end{subfigure}
    \end{framed}

    \vspace{-\baselineskip}
    
    \caption{The ChatGPT base \texttt{R} code (with one minor change) and its associated $\Bar{X}$ chart. 
    }
    \label{fig:baseR_and_plot}
\end{figure}

In our second question, we asked ChatGPT to ``rewrite the code above using base R only''. Figure \ref{fig:baseR_listing} presents the output from ChatGPT with our minor modification in the second line. If we were to ignore our change momentarily, one could observe that ChatGPT was able to successfully: (a) replace the \texttt{tidyverse}-heavy \texttt{group\_by}, \texttt{mutate}, and \texttt{ungroup} style into an appropriate \texttt{base R} coding style with the use of \texttt{rowMeans}, and \texttt{apply} (with a custom function to define the range instead of using the built-in \texttt{range} function); and (b) produce a \texttt{base R} equivalent plot (see Figure \ref{fig:baseR_plot}) to our earlier \texttt{ggplot2}-approach. 

ChatGPT's use of the \texttt{group = row\_number(data)} was incorrect for two main reasons. First, the \texttt{row\_number()} function comes from the \texttt{dplyr} package, which is not from \texttt{base R} and was not loaded in its generated prompt. Second, the use of \texttt{row\_number} in our earlier code was appropriate/correct due to the \texttt{group\_by()} operator. However, without grouping, the implementation is equivalent to the \texttt{base R: rank(ties.method = ``first'')}. The input to the rank function should be a vector; hence, the output from the ChatGPT code is a ranking of 80 values ($20 \times 4$, i.e., our 20 subgroups, each containing four observations). With this incorrect numbering/ranking and \texttt{R's} defaults of plotting data from left to right, the lines connecting the subgroups no longer capture the correct time order, and the resulting plot would be incorrect. Obviously, this error was mitigated by using \texttt{row.names()} in our modified code snippet, which produced the correct plot. Despite this minor error, we were impressed with ChatGPT's output to this prompt. 

\begin{figure}[htb!]
    \centering
    \begin{framed}
    \begin{subfigure}[b]{\textwidth}
        \begin{lstlisting}[basicstyle =\footnotesize, 
        language = Python,
        backgroundcolor = \color{light-gray},
        escapeinside={(*@}{@*)},
        keywords={read\_csv, import, as, index, assign, mean, max, min, plot, axhline, xlabel, ylabel, title, show},
        keywordstyle=\bfseries,
        otherkeywords={!,!=,~,$,*,\&,\%/\%,\%*\%,\%\%,<-,<<-,/, +, -},
        alsoother={.$},
        showstringspaces=false,
        breaklines=true]
import pandas as pd
import matplotlib.pyplot as plt

data = pd.read_csv(``data.csv'')
data[`group'] = data.index + 1
data_means = data.assign(
avg=data[[`obs_1', `obs_2', `obs_3', `obs_4']].mean(axis=1),
R=data[[`obs_1', `obs_2', `obs_3', `obs_4']].max(axis=1) - data[[`obs_1', `obs_2', `obs_3', `obs_4']].min(axis=1)
)

plt.plot(data_means[`group'], data_means[`avg'], `bo', markersize=2)
plt.plot(data_means[`group'], data_means[`avg'], `k-', linewidth=1)

plt.axhline(data_means[`avg'].mean(), color=`k', linestyle=`solid')
plt.axhline(data_means[`avg'].mean() + 0.729 * data_means[`R'].mean(), color=`k', linestyle=`dashed')
plt.axhline(data_means[`avg'].mean() - 0.729 * data_means[`R'].mean(), color=`k', linestyle=`dashed')
plt.xlabel(`Subgroup')
plt.ylabel(`Mean')
plt.title(`xbar Chart for data')
plt.show()

        \end{lstlisting}

        \vspace{-0.5\baselineskip}
        
        \caption{The Python code generated by ChatGPT in lieu of Figure \ref{fig:code_listing}.}
        \label{fig:python_listing}
    \end{subfigure}

    \vspace{0.75\baselineskip}
    
   \begin{subfigure}[b]{\textwidth}
        \centering
        \includegraphics[width=0.95\textwidth, trim = {0 0.13in 0 0.13in}, clip]{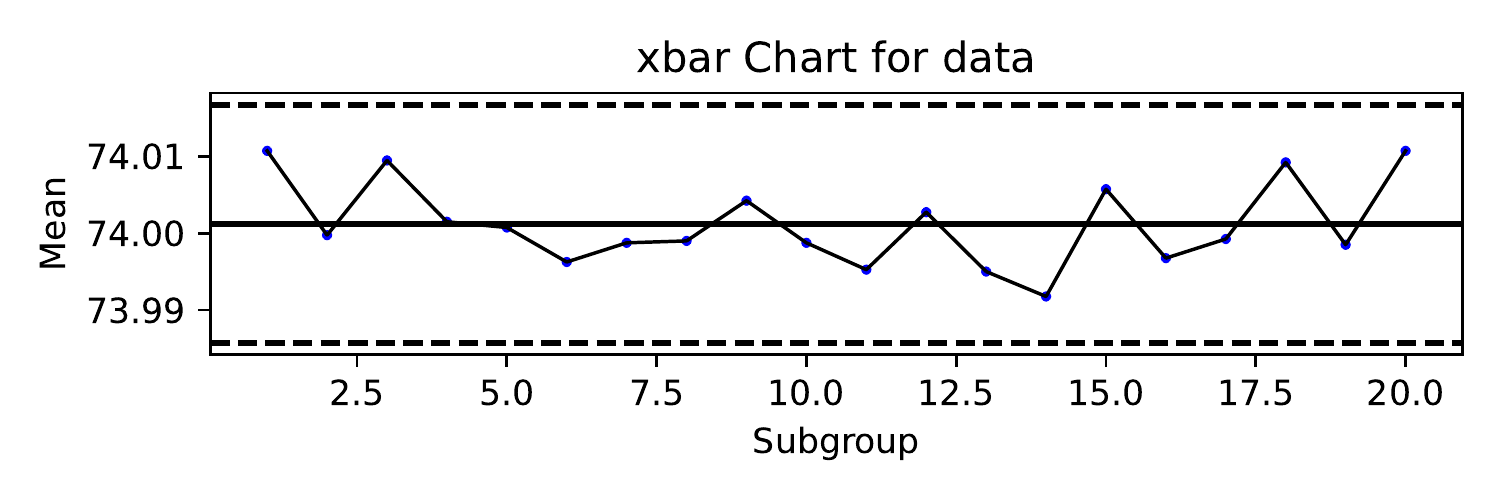}
        \caption{The correct $\Bar{X}$ chart produced with ChatGPT using the \texttt{pandas} and \texttt{matplotlib} libraries.}
        \label{fig:python_plot}
    \end{subfigure}
    \end{framed}

    \vspace{-\baselineskip}
    
    \caption{The ChatGPT generated \texttt{Python} code and its associated $\Bar{X}$ chart. 
    }
    \label{fig:python_and_plot}
\end{figure}

In our third question, we asked ChatGPT to rewrite the code from Figure \ref{fig:code_listing} in Python. To ensure that it is not affected by the slightly incorrect base R code, our prompt was ``Can you please rewrite the code below in Python?'' with the pasted code being the uncommented code version of Figure \ref{fig:code_listing}. The output from this prompt is shown in Figure \ref{fig:python_listing}, and the plot generated by running this code in a Python environment is shown in Figure \ref{fig:python_plot}. In this case, the generated code worked directly with no modifications. ChatGPT was able to identify the appropriate libraries/functions to rewrite our \texttt{R} code into \texttt{Python}. ChatGPT's success was likely possible since our original code was based on well-known packages that are well-documented online. Nevertheless, we think that ChatGPT's success here (even if it is limited to well-known packages) would reduce the burden of learning new programming languages and allow sharing (some) code quickly in multiple programming languages. 

\subsection{Explanation} 

\begin{figure}[htb!]
    \centering
    \includegraphics[width=0.95\textwidth, frame]{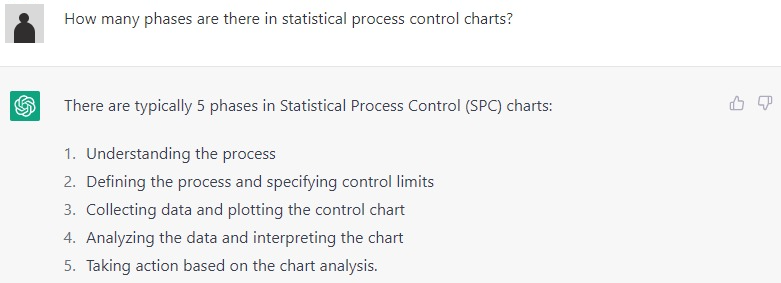}
    \caption{ChatGPT's statement that there are five phases in SPC control charts. 
    }
    \label{fig:five_phases}
\end{figure}

To examine ChatGPT's to provide learning-related explanations, we asked, ''How many phases are there in statistical process control charts?'' We intentionally worded the question similar to the first Practice-based explanation question (shown in Figure \ref{fig:phase1_2}), hoping for a two-phase answer. However, ChatGPT responded by stating that SPC charts have five phases (Figure \ref{fig:five_phases}). While we did not expect such an answer, and it is unlikely to be given by an SPC expert, it is not incorrect. One can argue that these are steps followed when implementing an SPC monitoring plan.

\begin{figure}[htb!]
    \centering
    \includegraphics[width=0.95\textwidth, frame]{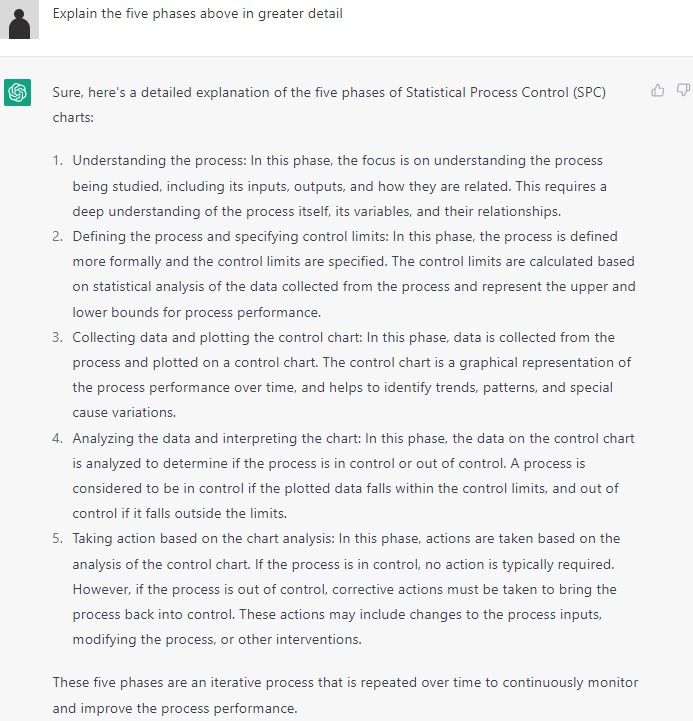}
    \caption{ChatGPT's ``in-depth'' explanation of the five phases in SPC control charts. 
    }
    \label{fig:five_phases_explanation}
\end{figure}

To ensure that we were not reading too much into ChatGPT's answer, we prompted it to ``explain the five phases above in greater detail.'' Figure \ref{fig:five_phases_explanation} depicts ChatGPT's explanation. To improve ChatGPT's explanation, we would have augmented its phases with the following (each bullet corresponds to one phase):
\begin{enumerate}[label = \arabic*., nosep]
    \item Included a discussion of measurement systems analysis to ensure that the gauges used in process measurement are repeatable and reproducible. The reader is referred to \cite{montgomery2020introduction} for a detailed introduction to measurement system analysis.
    \item ChatGPT's explanation of this step seems to be more centered around Phase 1 control chart applications, where the goals are to ensure that the process is in-control, estimate the process parameters and establish control limits (see our discussion in Section \ref{sec:practice_explanation}). To make the answer actionable, we would have expected a discussion of what to do if there is a plotted statistic outside the control chart in this phase. 
    \item This phase corresponds to Phase 2, where the process is monitored in real-time (or in fixed intervals if 100\% sampling is not employed). 
    \item[4-5.] Here, we do not have anything to add to the ChatGPT output. In our opinion, the answers for phases 4 and 5 are sufficient.
\end{enumerate}

\subsection{Knowledge Generation}

We asked ChatGPT to ``Generate a course syllabus for a statistical process control for undergraduate students. The students have already completed two statistics courses. The course is a three-credit hour course and is offered for 15 weeks.'' In our prompt, we ensure that ChatGPT knows the target audience, i.e., undergraduate students who have already completed two statistics courses and the course duration of 15 weeks (typical in the U.S. semester system). Its output, which was too long to include in one figure, consisted of several sections: 
\begin{enumerate}[label=(\alph*), nosep]
    \item \ul{introductory syllabus material:} which included the course title, prerequisites, duration, number of credit hours, and to be filled-in information about the instructor, their email, and office hours.
    \item \ul{course description:} ``This course provides an in-depth study of statistical process control (SPC) techniques and their applications in industry. Students will learn how to use statistical methods to monitor, control, and improve process quality in manufacturing and service industries. The course will cover both theoretical and practical aspects of SPC and include hands-on experience using statistical software.''
    \item \ul{learning outcomes:}
    \begin{itemize}[nosep]
        \item ``Describe the purpose and principles of statistical process control.''
        \item ``Choose and apply appropriate statistical methods for process control.''
        \item ``Use statistical software to perform process monitoring and control.''
        \item ``Interpret process control charts and identify special cause variation.''
        \item ``Evaluate process stability and determine appropriate process improvement actions.''
    \end{itemize}
    \item To-be-filled-in \ul{textbook}, \ul{grading}, and \ul{class policies}. 
    \item \ul{Schedule of topics}, which consisted of:
    \begin{itemize}[nosep]
        \item ``Week 1: Introduction to SPC''
        \item ``Week 2-3: Control charts for variables''
        \item ``Week 4-5: Control charts for attributes''
        \item ``Week 6-7: Process capability analysis''
        \item ``Week 8-9: Process improvement using SPC''
        \item ``Week 10-11: Design of experiments''
        \item ``Week 12-13: Multi-vari study'' {(\footnotesize{unclear if this meant multivariate SPC or the multi-vari chart})}
        \item ``Week 14-15: Group project and presentation'' 
    \end{itemize}
\end{enumerate}
We should note that the generated course description, learning outcomes, and schedule of topics were reasonable (can be adopted without major changes). We were impressed with this ChatGPT answer, which undoubtedly benefited from the many SPC syllabi found online and were likely included in its training. The interested reader is referred to our online supplementary materials for a link to the recording of ChatGPT's generated answer. 


\section{On the Use of ChatGPT in SPC Research}\label{sec:research}


\subsection{Code}

To examine ChatGPT's ability to generate code that can supplement potential SPC research projects, we provided ChatGPT with the following prompt ``calculate the zero-state ARL of an EWMA control chart with smoothing constant 0.1 and control limit factor 3 in the in-control case. The data is normally distributed. Use Monte Carlo simulation.'' ChatGPT responded with a Python-generated code, which we reproduce in Figure \ref{fig:python_listing2}. 

\begin{figure}[htb!]
    \centering
    \begin{framed}
   
    \begin{lstlisting}[basicstyle =\footnotesize, 
    language = Python,
    backgroundcolor = \color{light-gray},
    escapeinside={(*@}{@*)},
    keywords={read\_csv, import, as, index, assign, mean, max, min, plot, axhline, xlabel, ylabel, title, show, def, return, ewma, for, in, range, abs, if, break, print, simulate_ewma, std, zeros, random, normal, append},
    keywordstyle=\bfseries,
    otherkeywords={!,!=,~,$,*,\&,\%/\%,\%*\%,\%\%,<-,<<-,/, +, -},
    alsoother={.$},
    showstringspaces=false,
    breaklines=true]
import numpy as np
import matplotlib.pyplot as plt

def ewma(x, y, alpha):
    return (1 - alpha) * x + alpha * y

def simulate_ewma(n, alpha, k, num_simulations):
    arls = []
    for i in range(num_simulations):
        x = np.random.normal(0, 1, n)
        ewma_stat = np.zeros(n)
        ewma_stat[0] = x[0]
        for j in range(1, n):
            ewma_stat[j] = ewma(x[j], ewma_stat[j-1], alpha)
        control_limit = k * np.std(x)
        run_length = 0
        for j in range(n):
            if np.abs(ewma_stat[j]) > control_limit:
                run_length = j + 1
                break
        arls.append(run_length)
    return np.mean(arls)

n = 1000
alpha = 0.1
k = 3
num_simulations = 10000

arl = simulate_ewma(n, alpha, k, num_simulations)
print("Zero-state ARL:", arl)
    \end{lstlisting}
    \end{framed}

\vspace{-\baselineskip}
    
\caption{The ChatGPT-generated Python code in response to our request to calculate the zero-state ARL for the EWMA chart for normally distributed data via Monte Carlo simulation. 
}
\label{fig:python_listing2}
\end{figure}

Similar to the previous Python-generated code chunk in Figure \ref{fig:python_listing}, the above Python code runs without errors and produces a result. However, unlike the previous code chunk, this produced answer is incorrect. In particular, our testing of the \texttt{simulate\_ewma()} function resulted in values of 260.8, 255.8, 262.9, and 260.2. The correct result should be close to 842.1, 1035.1 (if we account for the skipped standardization), or 371 (if we corrected for interchanging \texttt{alpha} and \texttt{1-alpha}). In our diagnosis of the function, we identified two major flaws.  First, the run-lengths are truncated at \texttt{n = 1000}. In about 40\% of the Monte Carlo simulations, the truncation results in a zero run length making the final result small.  Second, the control limit incorporates the sample standard deviation of the underlying normal sample, which means that future/unseen data is used for establishing the control limit. We also observed minor flaws, including (a) the EWMA series is initialized with the first observation, which is unusual in the SPC literature (perhaps, ChatGPT confounded the initialization with the simple exponential smoother in time-series analysis), (b) interchanging \texttt{alpha} and \texttt{1-alpha}, and (c)  missing the standardization of the control limit, which is typically written in multiples of the EWMA standard deviation.

\begin{figure}[htb!]
    \centering
    \begin{framed}
    \begin{subfigure}[b]{\textwidth}
        \begin{lstlisting}[basicstyle =\footnotesize, 
        language = R,
        backgroundcolor = \color{light-gray},
        escapeinside={(*@}{@*)},
        keywords={function, sqrt, pnorm, outer, array, diag, solve, round},
        keywordstyle=\bfseries,
        otherkeywords={!,!=,~,$,*,\&,\%/\%,\%*\%,\%\%,<-,<<-,/, +, -},
        alsoother={.$},
        showstringspaces=false,
        breaklines=true]
ewma.arl <- function(lam, cE, mu, z0=0, r=50) {
  cE <- cE*sqrt(lam/(2-lam))
  ii <- (-r:r)
  w <- 2*cE/(2*r+1)
  qij <- function(i,j) pnorm((j*w-(1-lam)*i*w+w/2)/lam-mu) - 
    pnorm((j*w-(1-lam)*i*w-w/2)/lam-mu)
  Q <- outer(ii,ii,qij)
  one <- array(1,2*r+1)
  I <- diag(1,2*r+1)
  ARL <- solve(I-Q,one)
  arl <- ARL[r+round((z0+w/2)/w)]
  arl
}

        \end{lstlisting}
        \caption{The inputted \texttt{ewma.arl() R} function, which is to be explained by ChatGPT.}
        \label{fig:ewma_arl_fun}
    \end{subfigure}

    \vspace{\baselineskip}
    
   \begin{subfigure}[b]{\textwidth}
    \centering
    \includegraphics[width=0.95\textwidth]{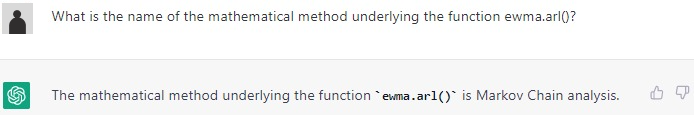}
    \caption{ChatGPT's response about the mathematical method underlying the function \texttt{ewma.arl().}}
    \label{fig:research_math}
    \end{subfigure}
    \end{framed}

    \vspace{-\baselineskip}
    
    \caption{Our input \texttt{ewma.arl()} function and ChatGPT's naming of the underling model. 
    }
    \label{fig:ewma_arl_with_explanation}
\end{figure}

In our second prompt, we asked to ``explain the following \texttt{R} function for approximating the ARL of a two-sided EWMA control chart for monitoring the mean of normally distributed random variables ....'' ChatGPT produced a verbose answer, which we do not reproduce here for space limitations (see our supplementary \textit{Online Materials} for a video recording of the output). However, we followed up with ``what is the name of the mathematical method underlying the function \texttt{ewma.arl()}?'' The corresponding answer was remarkably short and notably close to the complete answer ``Markov chain approximation''. The code and the corresponding ChatGPT response are presented in Figure \ref{fig:ewma_arl_with_explanation}.

\begin{figure}[htb!]
    \centering
    \includegraphics[width=0.92\textwidth, trim ={0 0 0 1.35in}, clip, frame]{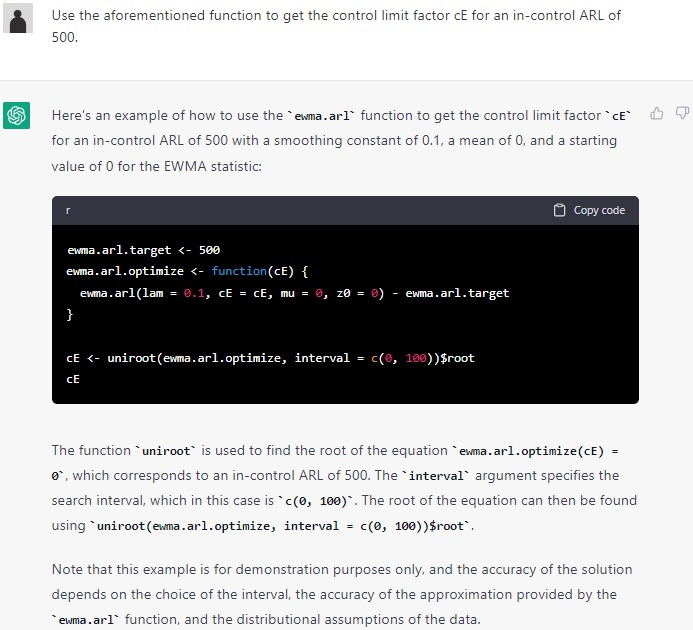}
    \caption{ChatGPT's code to get the control limit factor CE for an in-control ARL of 500.}
    \label{fig:research_code_control_limit}    
\end{figure}

As a final coding question, we asked ChatGPT to ``use the aforementioned function to get the control limit factor cE for an in-control ARL of 500.'' Note that this is the inverse task to that provided by our \texttt{ewma.arl()} function since we want to determine the control limit for a given ARL. ChatGPT's response is presented in Figure \ref{fig:research_code_control_limit}. While the solution steps are generally correct, the code delivers ``\texttt{f(lower) is NA}'' in its current form. The adverse interval choice in the \texttt{uniroot()} function
prevents solution generation. Both the lower and upper limits for the \texttt{interval} argument need to be changed. The lower limit has to be positive, otherwise the underlying function \texttt{ewma.arl()} fails to deliver a suitable result (we could obviously hard-code the ARL to equal 1 if the lower interval argument is zero). In our opinion, lifting the lower interval limit within the \texttt{uniroot()} function to a positive value is more appropriate. Furthermore, the upper limit has to be substantially smaller than 100. 
For \texttt{cE = 5}, the function \texttt{ewma.arl()}
would return an ARL $>$ 2,000,000.
Hence, an upper limit substantially larger than 5 is both unnecessary and creates numerical difficulties for \texttt{cE} values larger than 10 (with ARL values larger than $10^{16}$).
Setting  \texttt{interval = c(0.1,5)} results in 2.8156, which produces the correct outcome.

\subsection{Explanation}

\begin{figure}[htb!]
    \centering
    \includegraphics[width=0.75\textwidth, frame]{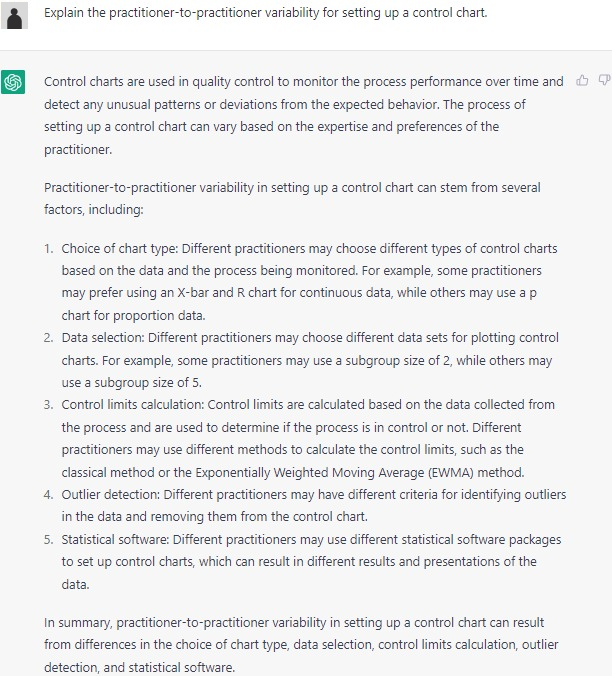}
    \caption{ChatGPT's imprecise explanation of the practitioner-to-practitioner variability. 
    }
    \label{fig:research_explanation}
\end{figure}

Here, we consider one of our explanation tasks and refer to the supplemental material for the other two. We asked ChatGPT to ``Explain the practitioner-to-practitioner variability for setting up a control chart''. We picked this task because it became a popular and handy notion for describing the uncertainty introduced by estimating population parameters for setting up the control limits of control charts \citep{zhang2013geometric,saleh2015difficulty,jardim2020two}. ChatGPT's response is presented in Figure \ref{fig:research_explanation}.

Similar to previous explanation tasks (in Sections \ref{sec:practice} and \ref{sec:learning}), the generated response is somewhat long and imprecise. Specifically, ChatGPT presented five factors, which share a common feature; all deal with differences on the \textit{method} level, i.e., chart type, subgroup design, techniques to calculate the limits, dealing with outliers, and choice of software. While we agree that these factors are important and will drive different results, ChatGPT's answer ignores the context for which the ``practitioner-to-practitioner'' variability is used in the SPC literature. In fact, the ``practitioner-to-practitioner'' variability refers to the variation that occurs with a fixed configuration of the five aforementioned factors, i.e., the variation results from multiple implementations of the same procedure on the same data-generating process. The interested reader is referred to \cite{zhang2013geometric} for a detailed introduction to ``practitioner-to-practitioner'' variability, and how the standard deviation of the average run length (SDARL) metric can be used to capture such variability.



\subsection{Knowledge Generation}
\begin{figure}[htb!]
    \centering
    \includegraphics[width=0.75\textwidth, frame]{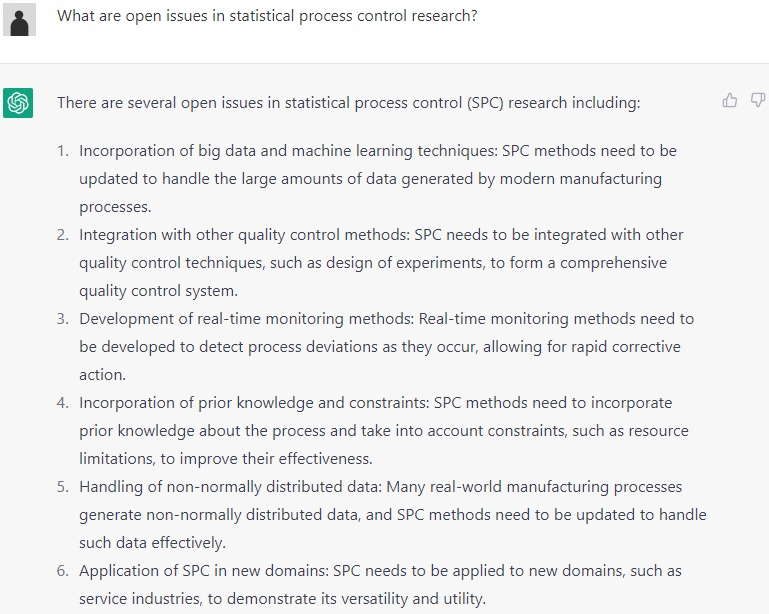}
    \caption{ChatGPT's identification of six open issues in SPC research. 
    }
    \label{fig:research_open_issues}
\end{figure}

Inspired by the TEDxBoston talk titled ``what we learned from 5 million books'' \citep{michel2011ted}, we were curious to see what is ChatGPT's answer to open issues in our field given that it likely ``read'' and ``can recall'' more SPC research papers than most (if not all) of us. Hence, we asked, ``What are open issues in statistical process control research?'' Its generated response, depicted in Figure \ref{fig:research_open_issues}, highlighted seven themes that included (a) incorporating big data and machine learning techniques, (b) online/real-time monitoring solutions where 100\% sampling is employed, (c) reducing the need for non-normality, and (d) applications to new domains.  

While the answer is reasonable, the real value, in our opinion, is the potential to use generative AI tools for research idea generation/validation. While this is likely to be only valid on a high level, this can be viewed through a similar lens to bibliometric approaches that identify literature clusters/themes based on keywords and citations \citep[e.g., see][]{megahed2019discussion}. Furthermore, the ideas can potentially be ``stale'' as the ChatGPT/GPT-3.5 model ``finished training in early 2022'' \citep{openai2022chatgpt}. It is important to note that the obsoleteness of responses may not be an issue in future LLM generations since there is a growing body of research examining the combination of external knowledge sources/APIs with LLMs \citep{karpas2022mrkl, schick2023toolformer}.

\section{Discussion and Conclusions}\label{sec:conc}

\subsection{Our Perspective on ChatGPT's (GPT-3.5) Ability to Augment SPC Practice, Learning, and Research}

In this expository paper, we attempted to examine whether generative AI models, such as ChatGPT, can inform SPC practice, learning, and research. To inform our understanding, our research team has engineered several prompts to capture some tasks that SPC practitioners, learners, and researchers may use to interact with ChatGPT. Our prompts within each grouping were divided into code, explanation, and knowledge generation subgroups to examine ChatGPT's utility/ability in performing these different tasks. From our results in Sections \ref{sec:practice}-\ref{sec:research}, we have several general observations on ChatGPT's ability to help the different SPC stakeholders. It is important to note that our observations are based on the \textit{ChatGPT Jan 30 Version (Free Research Preview)}, which we used in all of our interactions with ChatGPT.

First, one feature of generative AI models such as ChatGPT is that the answers to a given prompt are not repeated. We highlighted this feature in the introductory paragraph of Section \ref{sec:intro}. We were curious to see how this feature translates when interacting with ChatGPT. In our initial experimentation, our research team members used a sample of our prompts in the same order under similar conditions (i.e., we started a new conversation for each subtopic). The obtained outputs differed not only in terms of notation/wording but also in content. For example, our ``open issues in SPC research'' prompt generated responses that included \textit{design of experiments} and others that ignored this. Furthermore, the number of open issues differed by the response. In our opinion, this can be considered both a benefit and a limitation. On the one hand, the changing content can be viewed as a feature for prompts where there should not be one single answer (e.g., our open issues question or the notations used in the code). On the other hand, it is a possible deterrent in situations when a ``ground truth'' is needed. A detailed examination of singular-answer-prompts should be explored in future research.     

Second, we suspect many within the SPC community had not considered using AI to generate SPC code before ChatGPT's release. Our research background perhaps biases us, but we are excited about the idea of ``pair programming'' with a ChatGPT-like assistant. The advantages of ``pair programming'' (without the AI partner) are well-documented in computer science and software engineering \citep{begel2008pair}. Our assessment, based on ChatGPT January 30 Version, captures the following: (a) ChatGPT seems to be more proficient in \texttt{Python} than \texttt{R}, which we suspect is based on the fact that OpenAI does not seem to have trained its code completion module in the \texttt{R} programming language \citep{openai2022codex}; (b) its understanding of the arguments in some of the \texttt{R} functions are limited; (c) ChatGPT impressed us with its ability to explain functions, including capturing that our inputted \texttt{ewma.arl0()} function is a Markov Chain-based approach; and (d) ChatGPT can be useful in translating code from one programming language to another, especially when the functions are widely-used (which can be helpful in deployment SPC methods in a production environment). That being said, any code written by ChatGPT would need to be validated. We do not think it will eliminate the SPC expertise required in code development; however, it can augment and reduce the time taken to develop, test, and deploy code.   

Third, the current version of ChatGPT lacks the precision needed to correctly, consistently, and succinctly explain SPC concepts. In some instances, the generated output included the ``correct terms,'' but we judged the explanations to be of lower quality than expected from an SPC expert. In other instances, ChatGPT completely made-up explanations (e.g., in its description of the `bigfish' dataset). This issue is a known limitation of current generative AI models, including ChatGPT (see \citet{ji2022survey} for more details on \textit{hallucination in natural language generation}). To be fair, we did not attempt to optimize our prompts, especially in the practitioner and learning sections; however, our current prompts can be representative of what a practitioner and learner ask when they do not have additional context/details. 

Fourth, the knowledge creation prompts and their associated outputs highlight that ChatGPT can be useful as an idea generation or brainstorming tool. For example, the topics mentioned in response to our syllabus and open areas for research were generally reasonable. However, the weight given to each listed topic can differ by individual (especially since there is no single correct answer here).

To supplement our discussion and potentially biased perspectives, we asked ChatGPT to provide us with its view of its limitations and why we should not use it in SPC. Its response (which we document in Our Supplementary Material for space limitations) included: a lack of domain/specialized knowledge of the SPC field as it is trained on a wide range of topics, potential bias from its training data, and no capacity for performing data analysis. In our estimation, these limitations support our overall recommendation that ChatGPT output needs to be validated by multiple sources and experts. In the case of code, the validation involves two levels: (a) does it work? (i.e., runs with no errors), and (b) does it work as expected? (i.e., to be validated against known answers). For non-code-based explanations, we are not in favor of using ChatGPT. However, we think its use for knowledge creation is appropriate if it is treated as an initial brainstorming tool.

\subsection{Open Questions}

Generative AI is a rapidly evolving field with many open questions. In our opinion, our SPC community can contribute to advancing the field by investigating the following questions:
\begin{enumerate}[label=(\alph*), nosep]
    \item How can we effectively evaluate the quality of the generated outputs? This question is not only limited to text-to-text-based LLMs, but also to models that output audio, images, and videos. Furthermore, existing approaches to evaluating the quality provide ad-hoc mechanisms to account for the output variability resulting from a singular prompt. For example, in the code translation model space, the ``pass@k'' metric is often used, where a model is said to pass if one out of the $k$ generated output produces a correct translation \citep{weisz2023toward}. 
    
    \item Poor sampling and training of generative AI models can lead to biased predictions.  AI models can perpetuate and amplify biases when trained on biased data \citep{zhao2018bias,srinivasan2021biases,schwarz2021frequency}. In addition, generative AI models will collect and process possibly sensitive information, furthering our societal concerns on privacy and data security. We can see our community contributing in two complementary ways. First, our community can join/collaborate with generative AI companies to construct measures of training dataset bias and design sampling plans that eliminate/reduce the bias. In some domains, this is a relatively easy problem, e.g., in image generation, the gender distribution of the generated images can be used to detect dataset bias \citep{choi2020fair}. However, in other application domains, this problem is much more challenging. The second approach involves designing experiments post the public release of a generative AI model to examine potential biases in its outputs. 

    \item Some release versions of generative AI models have APIs that allow for custom-built applications. For example, GPT-3 can be tailored \citep{lim2021customizing} to SPC applications by providing it with SPC-based text for training (e.g., including \cite{montgomery2020introduction} with relevant journal papers and code repositories). Similarly, ChatGPT can be trained by including text in the chat window (where long documents can be broken down into smaller chunks using existing open-source tools). It is yet to be determined whether such an approach can better answer SPC-specific questions and prompts. 
\end{enumerate} 
In addition to these open questions, our community must be aware of many open ethical questions related to generative AI.  These include new guidelines excluding LLM from scientific authorship \citep{nature2023guidelines,thorp2023chatgpt,nature2023editorial}, discussion on ChatGPT's impact on science \citep{stokel2023chatgpt, nature2023tools}, and how LLM outputs can be watermarked \citep{kirchenbauer2023watermark} to prevent the misuse of AI-based output. We can also benefit from the examinations of LLMs' use in other disciplines \citep{dowling2023chatgpt, gilson2023does,biswas2023chatgpt,korinek2023language}. 

\subsection{Looking Forward}

Several phenomena will shape the future of generative AI models. First, LLMs have doubled in size every six months \citep{sevilla2022survey}. This has contributed to the rapid advancements in the abilities of these systems into usable application programming interfaces (APIs) that are widely used by the general public. Second, there is significant funding and investment interest in AI (see \citet{constantz2023big} for an in-depth analysis of how that translated into recent earning calls). Third, for-profit companies have primarily pushed these generative AI models (as indicated in Figure \ref{fig:LLM_models}). Since the data used in training these models are unknown to the general public, there will likely be some ethical/legal ramifications if copyrights were infringed. Fourth, we suspect that industry-specific generative AI models will be made available, providing tailored for specific application areas. Fifth, the AI race between several big companies will likely change our online experience. For example, Microsoft announced that it is incorporating a version of ChatGPT in its ``Bing'' search toolbar/site and ``Microsoft Edge'' browser (see \citet{mehdi2023} for the news release). 

The authors of this paper believe that generative AI and, specifically, LLMs will significantly change how we study, learn, and work. This is not the first technology to transform/disrupt society.  Our society has transformed with many technological disruptions, including steam-powered machines, computers, the internet, and advances in wireless/cellular/sensing technologies. With these disruptions, statistical process control has evolved as well:
\begin{enumerate}[label = (\alph*), nosep]
    \item \ul{Univariate methods}, such as the $\Bar{X}$ chart, were proposed to distinguish between common and special causes of variation. At the time, measurement was time-consuming, so the monitored statistic was univariate and sampled based on \textit{rational subgroups}; 
    
    \item With some advancement in computational methods, \ul{multivariate charts} were proposed to account for the correlation between several quality characteristics; 

    \item With the advancement in measurement and sensing technologies, it became possible to capture machine quality characteristics and a 100\% sampling of manufacturing products geometries via image-based sensing. This led to the development of control charts that can be used in \ul{time-series} \citep{alwan1988time}, \ul{profile} \citep{woodall2007current} and \ul{functional data} \citep{megahed2011review,wells2013statistical} monitoring applications;
    \item With the recent advancements in computational capabilities, our field has evolved to include \ul{statistical learning}-based control charting methods \citep{weese2016statistical}, \ul{big data} \cite{megahed2015statistical}, and \ul{predictive monitoring} strategies \citep{huberts2022multilevel}. 
\end{enumerate}
In addition, our community's productivity/efficiency has increased significantly with technological innovations in search engines, software (e.g., statistical analysis, version control, online meetings, and project management), and language models (which provide auto-complete suggestions in word processing, software, and browser applications).

At present, generative AI models like the one in ChatGPT are in their infancy.  However, the future of these has the potential to revolutionize many fields, including SPC. Currently, ChatGPT can, at a minimum, be a tool to improve our efficiency and reduce the time needed to complete some tasks (e.g., templates for documents/emails, code translation/explanation, and brainstorming). In the future, we suspect generative AI tools will be incorporated into many software tools. It is up to us as a community to examine how to use these innovations to further our practice, learning, and research agendas. \citet[p. 20]{box2012innovation} stated that ``quality and efficiency cannot compete against the right innovation ... [recounting] Deming’s story of the buggy whip manufacturer who had a highly efficient process with excellent quality but whose company collapsed because of the failure to foresee and adapt to the horseless carriage.'' With generative AI likely to stay, we need to be proactive and creative in examining how to use this technology to add value to the SPC profession, and how the SPC profession can help advance this technology.

\section*{Online Materials}

Our GitHub Repository, \url{https://github.com/fmegahed/llm_expository}, contains the code/data used in creating Figure \ref{fig:introductory}, screenshots of all our ChatGPT interactions (under the \texttt{figs} subfolder), code we used to assess ChatGPT's generated code for the Practice Section, and a Markdown containing videos of ChatGPT's answers to each prompt in Figure \ref{fig:study_design}. We provide these materials to document the outputs of our expository study given that the outputs from ChatGPT cannot be reproduced due to its inherent stochastic generation mechanism.  

\bibliographystyle{chicago}
\bibliography{refs}

\end{document}